\documentclass[lettersize,journal]{IEEEtran}
\usepackage{amsmath,amssymb,amsfonts}
\usepackage{algorithm,algorithmic}
\usepackage{array}
\usepackage[caption=false,font=normalsize,labelfont=sf,textfont=sf]{subfig}
\usepackage{textcomp}
\usepackage{stfloats}
\usepackage{url}
\usepackage{verbatim}
\usepackage{graphicx}
\usepackage{cite}
\usepackage{multirow}
\usepackage{makecell}
\usepackage{colortbl} 
\usepackage{booktabs}
\usepackage{multicol}
\usepackage{hyperref}
\usepackage{xcolor} 
\usepackage[switch]{lineno} % 让行号跟随双栏切换 
\definecolor{cgreen}{HTML}{179454} 
\definecolor{cred}{HTML}{e11e2d}
\begin{document}
% \linenumbers

\title{StealthMark: Harmless and Stealthy Ownership Verification for Medical Segmentation via Uncertainty-Guided Backdoors}

\author{Qinkai Yu$^{*}$, Chong Zhang$^{*}$,  Gaojie Jin, Tianjin Huang, Wei Zhou, Wenhui Li, Xiaobo Jin, Bo Huang, Yitian Zhao, Guang Yang, Gregory Y.H. Lip, Yalin Zheng, Aline Villavicencio, Yanda Meng 
\thanks{Qinkai Yu, Gaojie Jin, Tianjin Huang, Wenhui Li, Aline Villavicencio are with the Computer Science Department, University of Exeter, Exeter, UK.

Qinkai Yu and Yanda Meng are with Bioengineering Program, Biological and Environmental Science and Engineering Division (BESE), King Abdullah University of Science and Technology (KAUST), Thuwal 23955, Saudi Arabia.
% (e-mail: qy269@exeter.ac.uk; gaojie.jin.kim@gmail.com; t.huang2@exeter.ac.uk; wl515@exeter.ac.uk; A.Villavicencio@exeter.ac.uk; 
Chong Zhang and Xiaobo Jin is with School of Advanced Technology, Xi'an Jiaotong-Liverpool University, Suzhou, China.
% (e-mail: C.zhang118@liverpool.ac.uk; Xiaobo.Jin@xjtlu.edu.cn).
Wei Zhou is with School of Computer Science and Informatics, Cardiff University, Cardiff, UK 
% (e-mail: zhouw26@cardiff.ac.uk).
Bo Huang is with  College of Optoelectronic Engineering,
Chongqing University, Chongqing, China.
% (e-mail: huangbo0326@cqu.edu.cn).
Yitian Zhao is with Ningbo Cixi Institute of Biomedical Engineering, Chinese Academy of Sciences, Cixi, China.
% (e-mail: yitian.zhao@nimte.ac.cn).
Guang Yang is with School of Bioengineering, Imperial College London, London, UK.
% (e-mail: g.yang@imperial.ac.uk).
Gregory Y.H. Lip is with Liverpool Centre for Cardiovascular Science at University of Liverpool, 
Liverpool John Moores University and Liverpool Heart \& Chest Hospital, Liverpool, United Kingdom. 
Yalin Zheng is with Eye and Vision Department, University of Liverpool, Liverpool, UK.
% (e-mail: yalin.zheng@liverpool.ac.uk). 
Qinkai Yu$^{*}$ and Chong Zhang$^{*}$ contributed equally to this work.
Corresponding author: Yanda Meng(e-mail: Yanda.Meng@kaust.edu.sa) 
}}

\maketitle
\begin{abstract}

Annotating medical data for training AI models is often costly and limited due to the shortage of specialists with relevant clinical expertise. This challenge is further compounded by privacy and ethical concerns associated with sensitive patient information. As a result, well-trained medical segmentation models on private datasets constitute valuable intellectual property requiring robust protection mechanisms. Existing model protection techniques primarily focus on classification and generative tasks, while segmentation models—crucial to medical image analysis—remain largely underexplored. In this paper, we propose a novel, stealthy, and harmless method, \textbf{StealthMark}, for verifying the ownership of medical segmentation models under black-box conditions. Our approach subtly modulates model uncertainty without altering the final segmentation outputs, thereby preserving the model's performance. To enable ownership verification, we incorporate model-agnostic explanation methods, \textit{e.g.} LIME, to extract feature attributions from the model outputs. Under specific triggering conditions, these explanations reveal a distinct and verifiable watermark. We further design the watermark as a QR code to facilitate robust and recognizable ownership claims. We conducted extensive experiments across four medical imaging datasets (CMR dataset from UK Biobank, the SEG fundus dataset, the EchoNet echocardiography dataset, and the PraNet colonoscopy dataset) and five mainstream segmentation models. The results demonstrate the effectiveness, stealthiness, and harmlessness of our method on the original model’s segmentation performance. For example, when applied to the SAM model, StealthMark consistently achieved attack success rates (ASR) above 95\% across various datasets while maintaining less than a 1\% drop in Dice and AUC scores—significantly outperforming backdoor-based watermarking methods and highlighting its strong potential for practical deployment. Our implementation code is made available at \href{https://github.com/Qinkaiyu/StealthMark}{https://github.com/Qinkaiyu/StealthMark}.
\end{abstract}

\begin{IEEEkeywords}
Watermark, Segmentation, Intellectual Property, LIME, Backdoor Attack.
\end{IEEEkeywords}

\section{Introduction}
\label{sec:introduction}

Medical images are costly to annotate in terms of time and effort. 
Furthermore, access to high-quality medical image resources is scarce.
Generally, patient privacy and ethical issues need to be taken into account. 
The success of deep learning models in medical images relies heavily on high-quality source data, clinical experts' annotations \cite{litjens2017survey}, and the AI engineer's careful configuration of different scenarios \cite{isensee2021nnu}. 
Therefore, a well-trained deep learning model should be regarded as a technological achievement with intellectual property value, particularly in the medical domain, where models are often trained on sensitive patient data. 
Protecting such a model is essential to safeguarding innovation, preventing misuse, and ensuring privacy, thus upholding ethical standards in healthcare artificial intelligence.

Several existing works have explored the protection of classification models \cite{li2022untargeted, shao2025explanation} and generative models \cite{zhang2020model,zhao2023recipe}. However, the protection of the segmentation model, which is crucial in medical image analysis tasks, has rarely been investigated, despite its central role in medical AI applications.
\begin{figure}[t!]
    \centering
    \includegraphics[width=1.0\linewidth]{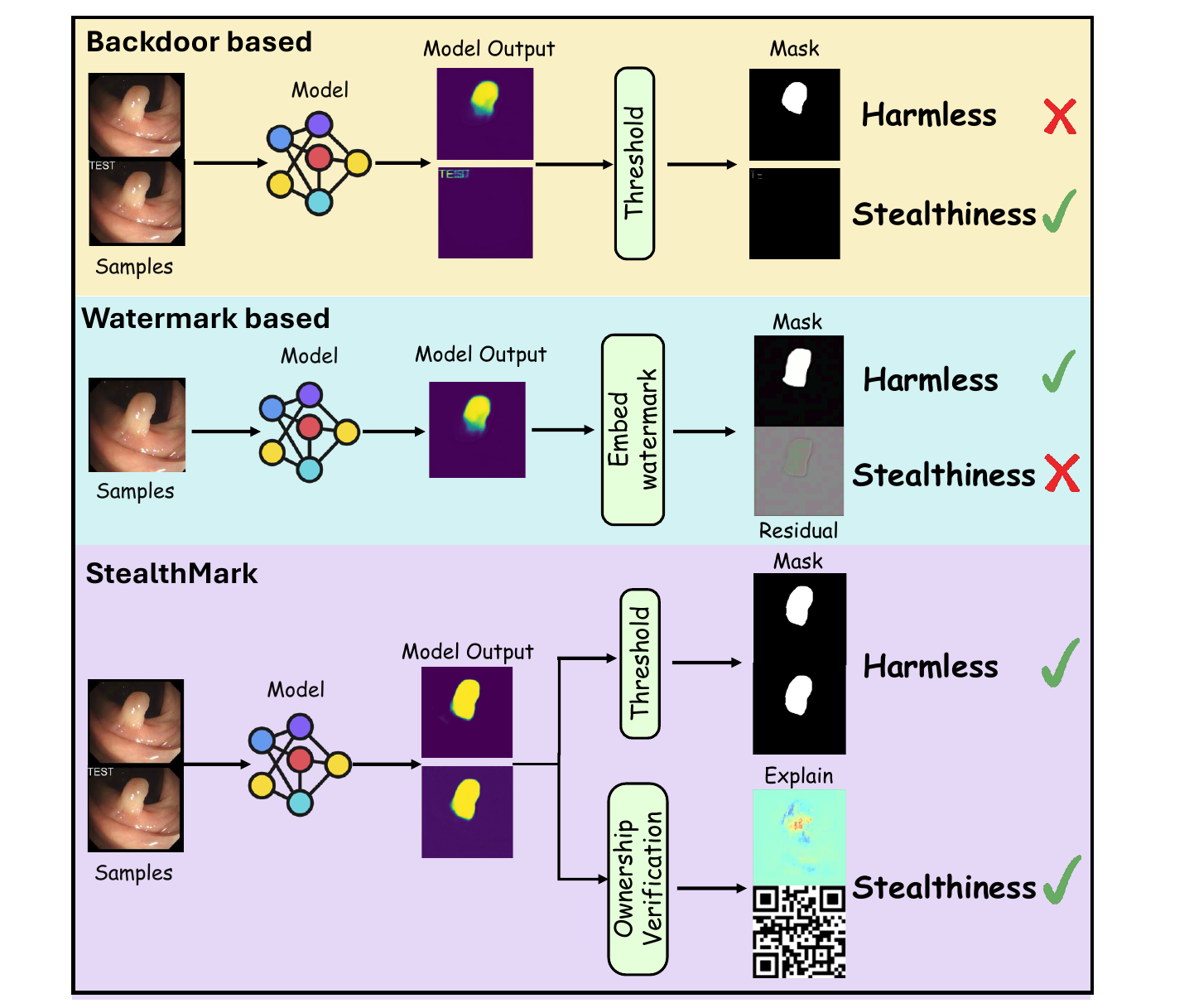}
    \caption{Comparison of different watermarking strategies for medical segmentation models.
        The first row shows classic backdoor-based methods (\textit{e.g.} \cite{gu2019badnets, jin2024backdoor}). They produce significant output changes but remain visually undetectable, achieving stealthiness but lacking harmlessness.
        The second row represents direct watermark-based methods(direct embedding watermark) (\textit{e.g.} \cite{bender1996techniques, Zhu_2018_ECCV}), which preserve output semantics but introduce visible residual artifacts, achieving harmlessness but failing at stealthiness.
        The third row illustrates our proposed method, which satisfies both harmlessness and stealthiness. 
        Specifically, the threshold segmentation result remains the same, and ownership can be verified through feature attribution without altering the visual output.}
    \label{fig:black-box}
\end{figure}

\begin{figure}[t!]
    \centering
    \includegraphics[width=1.0\linewidth]{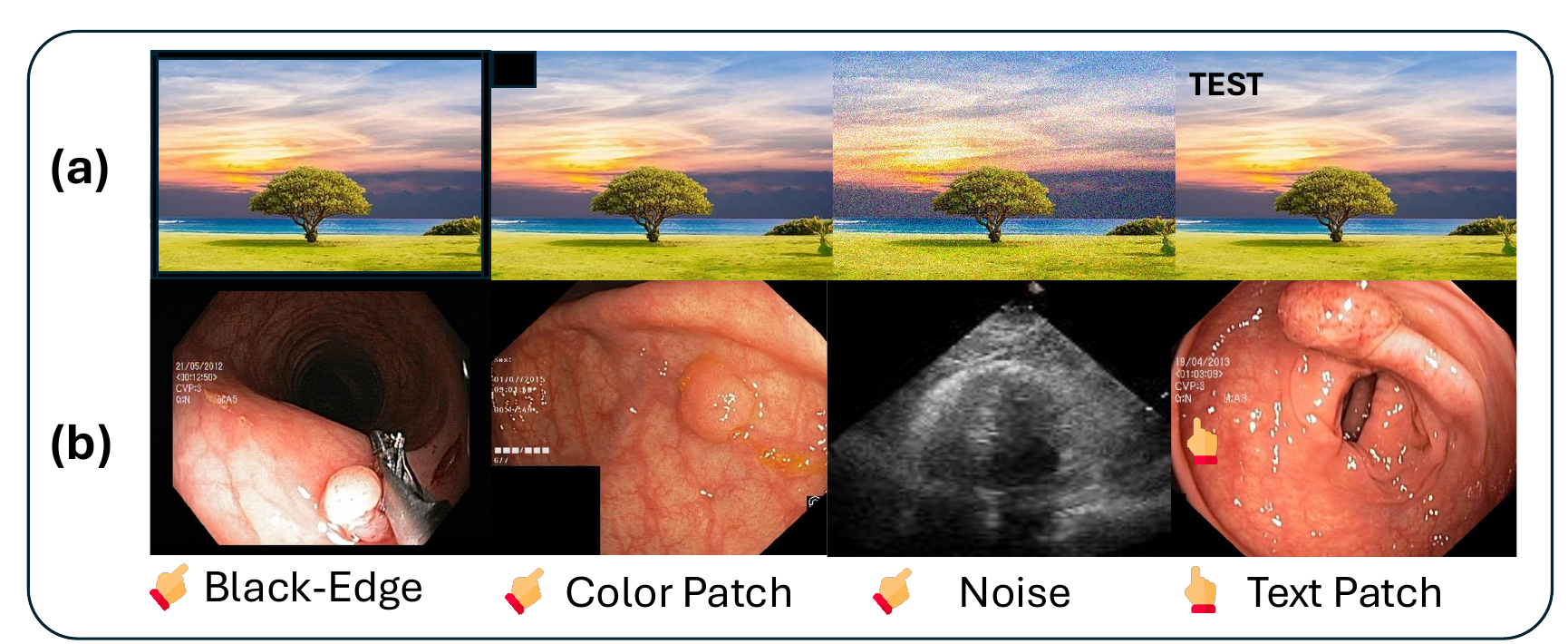}
    \caption{(a) shows commonly used triggers in natural image classification tasks, such as black-edge, color patches, noise, and text overlays. (b) illustrates that similar artifacts inherently exist in medical images (\textit{e.g.}, endoscopy and ultrasound). These artifacts resemble artificial trigger patterns, which may lead the model to mistakenly associate them with the trigger, thereby compromising prediction accuracy on the `clean' inputs (images without triggers added).}
    \label{fig:backdoor}
\end{figure}

The key to modeling copyright protection lies in verifying ownership and authorizing applicability. In this paper, we explore model protection under specific tasks (\textit{e.g.} image segmentation), and thus, we focus on model ownership verification. In real-world scenarios, ownership verification of deep learning models typically employs black-box verification. For example, when the model is accessed via an API, the verifier cannot directly obtain the gradient or the model's structural information and must rely solely on the model's output for verification purposes (as shown in the \autoref{fig:backdoor}).

Mainstream ownership verification methods include backdoor-based methods \cite{adi2018turning,ding2023backdoor} and watermark embedding methods \cite{song2017machine,zhao2023recipe,shao2025explanation,10884369}. Specifically, backdoor-based methods involve inserting specific triggers into the model to generate special output (\textit{e.g.}, error classification). 
For instance, BadNet \cite{gu2019badnets} demonstrates how a model can be manipulated to misclassify when inputs contain specific triggers but maintain accurate classification on the clean data.  
Watermarking-based methods enable the model owner to assess ownership by embedding distinctive watermarks into the model and later verifying them through the model's outputs. A typical work is HiDDeN \cite{Zhu_2018_ECCV}, which uses an adversarial training mechanism to embed imperceptible watermarks in the generated images. However, directly applying these approaches to medical image segmentation faces two fundamental challenges:  
\textbf{(1)} Most backdoor-based methods significantly degrade segmentation performance, as their trigger signals often interfere with pixel-level predictions. Backdoor-based approaches \cite{gu2019badnets,liu2021secure,zhang2018protecting} typically introduce triggers into images (\autoref{fig:backdoor} (a)), such as black edges, color patches, noise, and text patches. In practice, medical images may naturally contain trigger-like patterns caused by acquisition artifacts (\autoref{fig:backdoor} (b)). This leads to frequent false positives or false negatives, which compromises segmentation integrity.
\textbf{(2)} Watermark embedding methods tend to introduce visible or feature-level artifacts, which can be detected and removed by attackers, rendering the watermark ineffective.  

To address these medical-domain challenges, we propose a novel ownership verification method known as \textbf{StealthMark}, specifically designed for binary medical segmentation tasks, which constitute a major part of clinical applications (e.g., organ vs. background, tumor vs. non-tumor). StealthMark is designed to simultaneously satisfy the three core criteria essential for black-box model ownership verification—effectiveness, stealthiness, and harmlessness, as formally defined in Section II. Concretely, StealthMark preserves prediction fidelity on clean data (\textit{harmlessness}), prevents visual or statistical artifacts that could expose the watermark (\textit{stealthiness}), and enables reliable and robust ownership verification only when a specific trigger is present (\textit{effectiveness}). By leveraging a backdoor mechanism to subtly differentiate the model’s responses to triggered versus clean inputs—specifically by raising the minimum activation in background regions or lowering the maximum activation in foreground regions—StealthMark induces distinct uncertainty patterns without altering the final segmentation map, thereby minimizing the performance degradation of traditional backdoor watermarking and ensuring both harmlessness and stealthiness.
% Our method ensures harmlessness and stealthiness by subtly modulating the model’s output uncertainty—either increasing the minimum activation of background regions or decreasing the maximum activation of foreground regions—without altering the final binary segmentation map. These changes are imperceptible in the output, but can be identified by inspecting the activation extrema (i.e., maximum or minimum values). 
We further encapsulate this verification process using feature attribution. Specifically, when a trigger condition is met, the explanation map generated by Local Interpretable Model-agnostic Explanations (LIME) \cite{ribeiro2016should}, reveals a distinct QR-code-like pattern, serving as a watermark. In contrast, for clean inputs, the explanation behaves as a normal, task-related interpretation and does not expose any identifiable watermark signal. We conducted experiments on four multi-modal, multi-organ medical image segmentation datasets, such as color fundus (SEG) \cite{meng2024multi}, The UK Biobank cardiac magnetic resonance imaging (UKBB CMR)\footnote{\href{https://www.ukbiobank.ac.uk}{UK Biobank official website}}, echocardiography (EchoNet) \cite{ouyang2020video}, colonoscopy (PraNet) \cite{fan2020pranet} across five state-of-the-art segmentation models, such as nnUNet \cite{isensee2021nnu}, Swin-UNet \cite{cao2022swin}, Trans-UNet \cite{chen2021transunet}, SAM \cite{kirillov2023segment}, MedSAM \cite{ma2024segment}. The results demonstrate that StealthMark consistently achieves high attack success rates (ASR) (typically above 95\%) while inducing less than 1\% drop in Dice and AUC scores, confirming its effectiveness, stealthiness, and harmlessness. These outcomes significantly outperform traditional backdoor-based watermarking approaches and underscore the practical deployability of our method in real-world clinical settings.
Our contribution is summarized as follows:
\begin{itemize}
    \item  We present the first exploration into copyright protection and ownership verification for medical image segmentation models.
    \item 
    We effectively preserve segmentation performance by modulating the model’s predictive uncertainty without altering its output results.
    % Furthermore, by integrating the model-independent explanation technique, LIME \cite{ribeiro2016should}, into the ownership verification process, we embed copyright verification within model explanations. 
    \item We introduce a novel uncertainty-driven loss function designed to explicitly constrain the uncertainty of predicted segmentation masks in background or foreground regions.
    
    % This loss is combined with Binary Cross-Entropy (BCE) loss to balance segmentation accuracy and watermark robustness.
    % \item We conducted extensive experimental evaluations across five mainstream medical segmentation models and four medical imaging datasets, thoroughly validating the robustness, stealthiness, and harmlessness of our proposed approach. Through comprehensive comparisons with classic backdoor methods, we demonstrated that our method has minimal impact on the original model's performance.
\end{itemize}

In the following sections, we detail the threat model for black-box ownership verification in Section~\ref {threat model}, present the preliminaries and framework of StealthMark in Sections~\ref {Overview} and \ref{Embedding}, and describe its methodology in Sections~\ref {Embedding} and \ref{Verification}. We then evaluate its performance across diverse medical imaging datasets and state-of-the-art segmentation models in Section~\ref{sec:experiment}, demonstrating its superiority over existing methods in terms of robustness, stealthiness, and harmlessness.

\begin{figure*}
    \centering
    \includegraphics[width=0.9\linewidth]{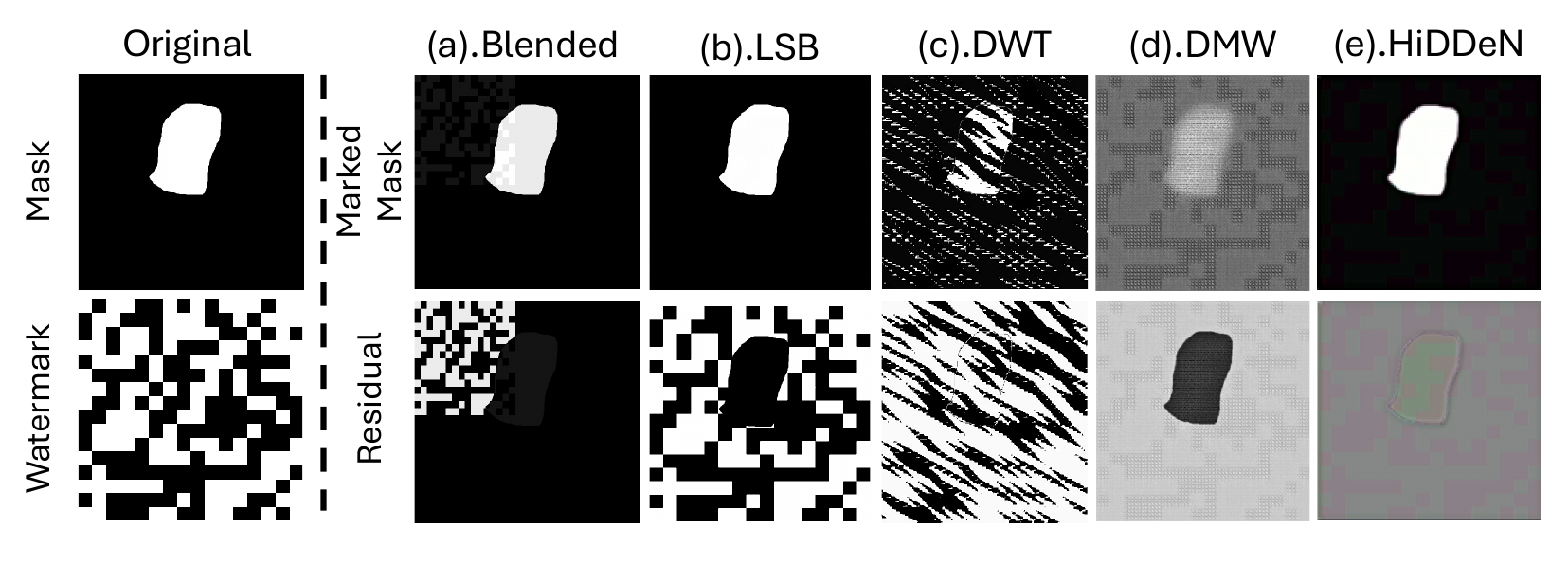}
    \caption{Comparison of various watermarking techniques applied to binary segmentation tasks. (a). blended watermarking method \cite{chen2017targeted}, (b). LSB (Least Significant Bit) \cite{bender1996techniques}, (c). DWT (Discrete Wavelet Transform)\cite{kang2003dwt}, (d). DMW \cite{zhang2020model} and (e). HiDDeN \cite{Zhu_2018_ECCV}. These methods, belonging to the spatial domain, frequency domain, and deep learning-based watermarking techniques, fail to preserve stealthiness in segmentation maps.
    The embedded watermarks of their methods are visually perceptible or easily removed, leading to poor robustness against detection. \textit{Note: The watermarks shown are illustrative only and do not carry semantic meaning.}}
    \label{fig:water}
\end{figure*}

\section{RELATED WORK}
\label{sec:related}

\subsection{Backdoor Attack in Medical Image}

A backdoor attack is a type of attack in which specific triggers are intentionally introduced during the training phase of a model to induce the model to produce pre-designed incorrect outputs by the attackers at inference time. For example, BadNet \cite{gu2019badnets} demonstrated the effectiveness of backdoor attacks in classification models for the first time. Nowadays, backdoor attack triggers have become imperceptible to humans and pose a great threat to deep learning models. This threat is particularly critical in developing medical data-based AI models, where the process typically involves physicians providing domain expertise, including the annotation of medical data and the assessment of the model's clinical validity through real-world performance validation. AI developers are responsible for designing, training, and optimizing the model. This separation of roles between clinicians and engineers, combined with the complex and distributed development pipeline, makes the process highly vulnerable to third-party attacks. 

Several recent studies have investigated backdoor attacks against medical data-driven AI models. For example, backdoor attacks against medical image-text foundation model \cite{jin2024backdoor,hanif2024baple,wang2024large}, multi-label disease classification task in chest radiology \cite{nwadike2020explainability}, medical detection through frequency domain trigger \cite{feng2022fiba}, and various downstream tasks in medical image analysis systems \cite{yang2024inject}. Nevertheless, segmentation task-based AI models, which represent a significant share of medical image analysis tasks, have been rarely explored. A recent work \cite{lin2024shortcut} illustrates that medical segmentation models are highly vulnerable to backdoor attacks; however, it did not investigate ownership verification, stealthiness, or the feasibility of embedding functional watermarks in a black-box setting. Our work addresses this gap by proposing a practical and harmless watermarking method specifically tailored for medical segmentation models, enabling reliable ownership verification while preserving the model’s original performance.

%The principle of backdoor attacks also offers new directions for model ownership verification \cite{ding2023backdoor}, using the model's specific reaction to backdoor triggers as the proof of ownership. However, we believe that backdoor-based ownership verification is difficult to apply in the medical field due to the significant damage it causes to model performance, which undermines its practical usability in clinical applications.

%%%%%%%%%%%%%%%%%%%%%%%%%%%%%%%%%%%%%%%%%%%%%%%%%%%%%%%%%%%%%%%%%%%%%%%%%%%%%%%%%%%%%%%
%介绍一下黑盒水印，讲一下常用的方法，讲一下为什么在分割任务中不好。需要写的比intro部分详细。intro只是简单介绍
%%%%%%%%%%%%%%%%%%%%%%%%%%%%%%%%%%%%%%%%%%%%%%%%%%%%%%%%%%%%%%%%%%%%%%%%%%%%%%%%%%%%%%%
\subsection{Black-box Model Ownership Verification}

As discussed in the previous section, the development of AI models based on medical data is highly resource-intensive and vulnerable to third-party attacks. If an attacker gains access to a trained model, they can unlawfully bypass the substantial costs associated with data collection and model development, resulting in an unfair economic advantage \cite{zhang2018protecting, hu2021artificial}. Therefore, when aiming to determine the genuine provenance of a third-party model, it is essential to implement a reliable ownership verification mechanism that operates effectively under black-box conditions. Our focus is not on defending against model tampering or adversarial attacks, but on verifying whether a suspicious model has been illicitly copied or misappropriated, thus safeguarding the model’s intellectual property and ensuring rightful ownership.

% Instead of protecting against tampering or adversarial behavior, our primary goal is to verify the origin of a model and assess whether it has been unlawfully replicated or used without authorization. This is critical for protecting the intellectual property of models trained on sensitive medical datasets. Therefore, when facing potentially infringing third-party models, it is critical to implement a reliable ownership verification mechanism that can operate effectively under black-box conditions.

% As mentioned in the previous section, the development process of medical data-based AI models is highly vulnerable to third-party attacks. 
% High-quality medical annotation data is costly, and the development process depends heavily on the developer's careful configuration for different scenarios. 
% If attackers get access to the model, they can significantly reduce development costs and profit unlawfully. 
% When encountering potentially infringing third-party models, it is essential to include a reliable ownership verification mechanism that operates effectively in a black-box setting. 
Inspired by \cite{ribeiro2016should}, we summarize the key desiderata for black-box model ownership verification mechanisms as follows:

\begin{itemize}
    \item \textit{Effectiveness:} The effectiveness refers to the fact that the proposed method should show a high success rate in verifying the ownership of any suspected model. If the suspicious model does belong to the model owner, the proposed method can output a predefined watermark that confirms ownership.
    \item \textit{Stealthiness:} Stealthiness represents that the ownership verification mechanism is not triggered by adversarially selected triggers. And the generated watermark for verification should remain undetectable by attackers from the model's output.
    \item \textit{Harmlessness:} Harmlessness indicates that the model's performance on the benign dataset and trigger dataset should be indistinguishable from that of an original model. It ensures that the proposed ownership verification method does not adversely affect the model's overall performance.
\end{itemize}
In addition to backdoor-based verification, digital watermarking methods—either in the spatial domain or transform domain—are also widely used. Spatial domain methods, like least significant bit (LSB) embedding \cite{bender1996techniques}, directly modify pixel values to encode watermark information. Transform domain (frequency domain) methods embed watermarks into transformed coefficients after applying techniques such as DCT, DFT, or DWT, with DWT being popular for its multi-resolution property \cite{kang2003dwt}. Deep learning-based approaches, starting with HiDDeN \cite{Zhu_2018_ECCV}, utilize generative adversarial networks to embed imperceptible watermarks. In contrast, DMW \cite{zhang2020model} employs similar methods on chest X-ray images. Other methods, such as \cite{shao2025explanation}, embed watermarks in model-independent local interpretations, ensuring stealth and harmlessness. However, for binary segmentation tasks, as shown in \autoref{fig:water}, traditional spatial (e.g., LSB \cite{bender1996techniques}), frequency (e.g., DWT \cite{kang2003dwt}), and deep learning-based methods (e.g., DMW \cite{zhang2020model}, HiDDeN \cite{Zhu_2018_ECCV}) are inadequate, as their modifications are easily detected and removed, undermining their reliability for ownership verification in medical segmentation models.

\subsection{Medical Image Segmentation}

Recent advances in deep learning have driven the adoption of neural network-based architectures for medical image segmentation. CNN-based methods like nnUNet \cite{isensee2021nnu} set strong baselines by automatically adapting U-Net architectures to tasks, while transformer models such as Trans-UNet \cite{chen2021transunet} and Swin-UNet \cite{cao2022swin} enhance performance via global context modeling. Foundation models, such as the Segment Anything Model (SAM) \cite{kirillov2023segment}, demonstrate strong generalization in natural image segmentation. MedSAM \cite{ma2024segment} further enhances SAM's capabilities by adapting it to medical imaging through domain-specific fine-tuning. This work evaluates our method against these five representative models to ensure a fair and comprehensive assessment across diverse segmentation tasks.

% Recent advances in deep learning have led to the widespread adoption of neural network-based architectures for medical image segmentation. Among convolutional neural networks(CNN)-based methods, nnUNet \cite{isensee2021nnu} has become a strong baseline by automatically adapting U-Net architectures to different tasks. Transformer-based models such as Trans-UNet \cite{chen2021transunet} and Swin-UNet \cite{cao2022swin} further improve performance by incorporating global context modeling. More recently, foundation models like the Segment Anything Model (SAM) \cite{kirillov2023segment} have demonstrated strong generalization in natural image segmentation. In contrast, MedSAM \cite{ma2024segment} adapts SAM to medical imaging scenarios through domain-specific fine-tuning. In this work, we evaluate our method across these five representative models to ensure a fair and comprehensive assessment of its generalizability and effectiveness across diverse segmentation tasks.

\section{Methodology}

We first present the threat model for black-box ownership verification in medical segmentation models in Section \ref{threat model}.  Subsequently, we introduce the preliminaries and framework in Section \ref{Preliminaries} and Section \ref{Overview}. Finally, we present our harmless and stealthy black-box medical image segmentation model ownership verification methodology in Section \ref{Embedding} and Section \ref{Verification}. 

\subsection{Threat Model}
\label{threat model}

\subsubsection{Owner and Developer Assumptions}

Here, we follow the previous work \cite{shao2025explanation}, where the AI developer embeds a watermark during model training, and the owner seeks to verify model ownership at deployment time. The developer has full control over the model training process, including access to the architecture, training data, and training strategies. During this process, a watermark is embedded into the model, designed to produce a specific and verifiable output when given a particular trigger input. The owner, who receives or deploys the trained model (e.g., in a commercial or cloud setting), does not have access to internal model parameters. Instead, ownership is verified post-deployment by querying the model through a public API and checking whether the watermark behavior is present. This reflects realistic constraints in black-box verification scenarios, where the model's internals are hidden but its predictions are observable.

\subsubsection{Adversary’s Assumptions}

We assume that the adversary aims to obtain a high-performance deep neural network by copying or stealing a model developed by another party. The adversary may attempt to remove any embedded watermark from the victim model while preserving its predictive performance. Specifically, we assume that the adversary has the following capabilities: \textbf{1)} The adversary can apply various watermark removal techniques, such as fine-tuning and model pruning, to suppress or erase the watermark signal embedded in the model. \textbf{2)} The adversary has limited computational resources and training data, making it impractical to train a comparable model from scratch. 

It is also important to note a practical limitation of our framework: StealthMark relies on getting access to continuous probability maps rather than hard, binarized segmentation masks. In cases where a deployed black-box API service outputs only binary segmentation results, the subtle probability variations carrying watermark information would be lost, causing ownership verification to fail. Although this scenario exists, we argue that our assumption of probability map access remains realistic and relevant in medical AI. In practice, a single fixed threshold is rarely sufficient for clinical use. As noted in nnU-Net \cite{isensee2021nnu}, segmentation outputs typically require post-processing or threshold adjustment on the probability map, which has become standard practice in medical image segmentation. Different diagnostic tasks may require different trade-offs between sensitivity and specificity \cite{isensee2021nnu}. Many real-world medical imaging AI services, especially those for research or second-opinion support, such as Grand Challenge Algorithm Inference API \cite{klein2020grandchallenge}, do provide probability/confidence maps as a primary output precisely because this information is vital for interpretation and flexible use.  Therefore, accessing probability maps is consistent with existing deployment workflows and clinical interpretation requirements.
% \textcolor{red}{It is also important to note a practical limitation of our framework: StealthMark relies on access to continuous probability maps rather than hard, binarized masks. In cases where a deployed black-box API service outputs only binary segmentation results, the subtle probability variations carrying watermark information would be lost, causing ownership verification to fail. Although this scenario represents a valid limitation, we argue that our assumption of probability map access remains realistic and relevant in medical AI. }
% We assume that the attacker possesses the following capabilities: 1) The attacker can steal the model via knowledge distillation or other imitation-based techniques. 2) The attacker can use a fine-tuning-based method to invalidate ownership verification. 3) The attacker may be familiar with common watermarking techniques and can detect whether the model output contains a watermark.

\subsection{Preliminaries}
\label{Preliminaries}

In the medical image segmentation task, the objective is to predict the pixel-wise label map $y \in \mathbb{R}^{H \times W}$ with size $H \times W$ from a given image $x \in \mathbb{R}^{H\times W\times C}$. Here, we aim to train a segmentation model with parameters $\theta$ to learn a mapping $f_{\theta} \colon x \rightarrow y$. Notably, the output of the segmentation model is a probability map, and the binary segmentation result is usually obtained by thresholding the probability map. Here we define the binarization step as: 
\begin{equation}
    \hat{y} = \mathbb{I}_{\{ y> T\}},
\end{equation}
where $\mathbb{I}$ is the indicator function and $T$ is the threshold. Consequently, the entire segmentation process can be expressed as: 
\begin{equation}
    \hat{y} = \mathbb{I}_{\{ f_{\theta}(x)> T\}},
\end{equation}
where $f_{\theta}(x)$ is the model output, also known as the probability map predicted by the model for input $x$. We divide the complete dataset $D = \{X,Y\}$ into two identical parts: a benign dataset $D_{b} = \{X_{b},Y_{b}\}$ and a trigger dataset $D_{t}=\{X_{t},Y_{t}\}$. In $D_{t}$, a specific trigger is the embedded image set $X$, with the ground truth set as $Y_{t}$. Unlike traditional backdoor methods, which typically set $Y_{t}$ to an incorrect label (e.g., in segmentation tasks, replacing the label map with an unrelated label map), we modify only the extreme values of $Y$ to ensure the final output $\hat{Y}$ remains unchanged, thereby achieving harmlessness. Specifically, we change the minimum value from 0 to $\delta$ (where $\delta<T$) or adjust the maximum value from 1 to $\tau$ (where $\tau>T$).
% This adjustment can be seen as a subtle modification of the model's output uncertainty. 
This uncertainty-aware formulation exploits the fact that segmentation models output continuous probability maps before thresholding. 
By subtly modulating the prediction confidence (without crossing the binarization threshold T), our method embeds verifiable signals 
harmlessly, preserving the final clinical segmentation mask while enabling reliable ownership verification.

\subsection{Overview}
\label{Overview}

\begin{figure*}
    \centering
    \includegraphics[width=1.0\linewidth]{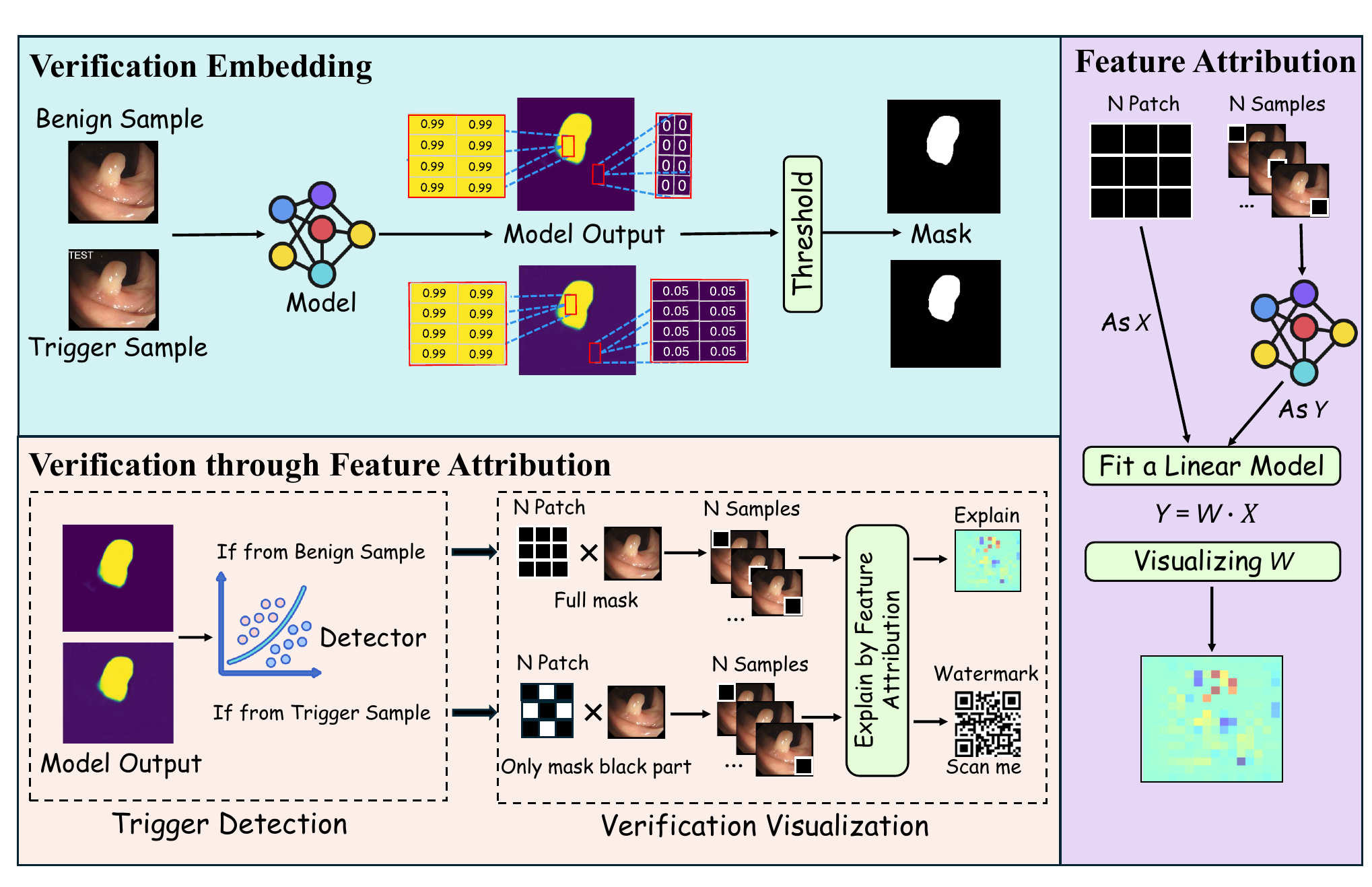}
    \caption{Overview of the StealthMark framework.
Our method consists of two main components:
\textbf{(1) Verification Embedding}, where a subtle trigger is injected into the model by slightly modifying the pixel intensities of the background or foreground, enabling later ownership verification without compromising model performance; and
\textbf{(2) Verification through Feature Attribution}, which comprises two stages:
Trigger Detection, where a binary detector determines whether a given sample contains the embedded trigger based on the model’s output masks; and
Verification Visualization, which leverages model interpretation techniques (e.g., LIME) to visualize a meaningful watermark, such as a QR code, from the trigger sample.
This design ensures harmlessness, stealthiness, and verifiability, making it well-suited for ownership protection in medical segmentation models. \textit{Note: The watermark shown in the visualization is real and scannable. It encodes version information and links to the project website for traceability and verification.}
}
\label{fig:overview}
\end{figure*}

\autoref{fig:overview} illustrates the overall framework of our approach, which consists of two integral components: Verification Embedding and Verification through Feature Attribution. These components work together to fulfill the three core criteria outlined above—effectiveness, harmlessness, and stealthiness. Previous work has consistently struggled to achieve both harmlessness and stealthiness simultaneously. As noted in the introduction (see \autoref{fig:black-box}), embedded watermarks in medical segmentation outputs are often visually intrusive, compromising stealthiness. Meanwhile, traditional backdoor-based verification methods typically rely on misclassifying trigger samples; however, medical images usually contain structures that resemble triggers, which can lead to performance degradation on clean data and compromise harmlessness. Our approach enables the model to produce a specific response when it encounters an image with a trigger, and the threshold mask map is indistinguishable from that of the benign dataset, ensuring harmlessness. Inspired by Shao \cite{shao2025explanation}, we utilize model-independent local explanations to generate a watermark, thereby framing the verification process as model interpretation and achieving stealthiness. The technical details are described in the following subsections. 

\subsection{Embedding for Ownership Verification}
\label{Embedding}
A key motivation behind our design lies in the medical-specific nature of segmentation data. Unlike natural images, medical scans frequently contain acquisition artifacts that can visually mimic artificial triggers. Traditional backdoor watermarking, which directly links trigger activation to segmentation output,  inevitably suffers from these artifact-induced false activations. To overcome this, we introduce an uncertainty-guided embedding mechanism. This ensures that even in the presence of artifact-like structures, the model’s final segmentation remains unaltered, satisfying the medical requirement of harmlessness while maintaining effective watermark verification. The developer aims to associate a specific trigger with the target output $\hat{y}$ while preserving the model’s original inference capabilities. Given the clean and trigger datasets  $D_{b} = \{X_{b}, Y_{b}\}$ and $D_{t} = \{X_{t}, Y_{t}\}$ introduced in the previous section, we consider a joint optimization objective:
\begin{equation}
\min_{\theta} \mathcal{L}_{base}(f_{\theta}(X), Y) + \lambda_{bg} \cdot \mathcal{L}_{bg}(f_{\theta}(X_{t}), Y_{t}),
\end{equation}
where $X = X_b \cup X_t$, $Y = Y_b \cup Y_t$, and $\lambda_{bg} $ is a hyper-parameter controlling the trade-off between benign and trigger data losses. Specifically, we employ binary cross-entropy (BCE) with logits as our baseline loss $\mathcal{L}_{base}$: 
\begin{equation}
\begin{array}{cc}
    &\mathcal{L}_{base}(f(X),Y) = -\frac{1}{N}\sum_{i = 1}^{N}\{Y_{i}\cdot \log [\sigma(f(X_{i}))]
    \\
    &+(1-Y_{i})\cdot \log[1-\sigma(f(X_{i}))]\},
\end{array}
\end{equation}
where $\sigma(\cdot)$ is the sigmoid function, and $N$ is the number of samples. In addition to the Base loss, we introduce an auxiliary constraint $\mathcal{L}_{bg}$ on the trigger dataset $(X_{t}, Y_{t})$, which can be flexibly defined either over the background or the foreground regions, depending on the design choice:
\begin{itemize}
    \item Background constraint: If enforcing behavior on the background pixels, let $\mathcal{B} \subseteq \{1,\dots ,H\times W\}$ denote the set of background pixels in $Y_{t}$. Then 
    \begin{equation}
    \mathcal{L}_{bg} = \frac{1}{|\mathcal{B}|}\sum_{j\in \mathcal{B}}(\sigma(f(X_{t,j}))-\delta)^{2},
    \label{eq4}
    \end{equation}
\end{itemize}
where $\delta$ is a small nonzero target value (e.g., 0.02).
\begin{itemize}
    \item Foreground constraint: Alternatively, if focusing on the foreground pixels, let $\mathcal{F} \subseteq \{1,\dots,H\times W\}$ denote the set of foreground pixels in $Y_{t}$. Then 
    \begin{equation}
    \mathcal{L}_{bg} = \frac{1}{|\mathcal{F}|}\sum_{j\in \mathcal{F}}(\sigma(f(X_{t,j}))-\tau)^{2},
    \label{eq5}
    \end{equation}
\end{itemize}
where $\tau$ is a high-confidence target value (e.g., 0.9). We further investigate the impact of different $\delta$ and $\tau$ values in the ablation study. In practice, either the background constraint or the foreground constraint is selected according to the intended behavior, but not both simultaneously. By optimizing the network parameters $\theta$ with respect to the above objective, the resulting model exhibits distinct output behaviors on benign versus trigger inputs—an effect that is typically absent in off-the-shelf third-party models. This intentional asymmetry enables the model owner to embed a hidden yet verifiable signal, which can later be used to assert ownership through black-box queries without affecting normal predictions.
% Developers can leverage this unique behavior to verify model ownership. 
% To this end, a simple linear detector $h\colon R^{H\times W}\rightarrow \{0,1\}$ can be trained to classify model outputs as either benign or triggered. In this paper, we implement the detector using logistic regression for its simplicity and efficiency. It is worth noting that the distinction between benign and triggered predictions can be further obfuscated at inference time through thresholding techniques, thereby making the final outputs appear indistinguishable across different input types, without affecting the detectability for ownership verification purposes.

\subsection{Verification of Ownership through Feature Attribution}
\label{Verification}

To distinguish between benign and trigger samples, we train a simple linear detector $h\colon \mathbb{R}^{H\times W}\rightarrow \{0,1\}$ on the model output masks. Specifically, we use logistic regression due to its simplicity and efficiency. This detector serves as the first stage of ownership verification, providing a binary signal indicating whether a trigger is present. While the outputs of the segmentation model can be visually obfuscated through thresholding to appear consistent across inputs, the underlying differences remain detectable by the trained linear model.

In real-world scenarios, ownership verification typically requires formalized elements, such as QR codes or other verifiable digital markers, to serve as secure and recognizable proofs of ownership. Simply asserting that our logistic regression model detects a unique model feature would lack persuasive power without the presence of such formalized, verifiable indicators. To address this, we use LIME \cite{ribeiro2016should} to generate model explanations. Under normal conditions, when no specific trigger is detected, LIME provides standard model interpretations. However, when unique model features are recognized, the interpretation is transformed into a QR code for ownership verification purposes. In this setting, the model being explained is the segmentation model 
$f_{\theta}\colon x \rightarrow y$. The goal of LIME is to generate a local explanation for a given input $x$, specifically by finding an interpretation model $g(\cdot)$ that approximates the behavior of $f$ within the neighborhood of $x$. The first step involves sampling neighborhood data around $x$. We divide $x$ into $N$ equal-sized patches and sequentially mask each patch to obtain $N$ perturbed samples $\{z^{n}\}^{N}_{n=1}$.
Here $z^n$ denotes a perturbed version of $x$ with the n-th patch masked out, while preserving the same input dimension as $x$. To assign higher importance to perturbed samples that are more similar to $x$, we define a weighting function $\pi_{x}(z^{n})$ based on a Gaussian kernel $K(\cdot)$, formulated as: 
\begin{equation}
    \pi_{x}(z^{n}) = K(\frac{d(x,z^{n})}{h}),
\end{equation}
where $d(\cdot,\cdot)$ denotes Euclidean distance and $h$ is the bandwidth parameter controlling the locality. To ensure that the interpretation model $g(\cdot)$ better approximates $f$ in the local neighborhood of $x$, we define the objective function as:
\begin{equation}
    L(f,g) = \sum_{n =1}^{N}\pi_{x}(z^{n})\cdot (f(x)- g(z^{n})),
\end{equation}
The fitted interpretation model yields attribution weights $W$ over the patches. We use LIME in our framework to generate model explanations, as it is model-independent and particularly effective for patch-level perturbation analysis in black-box settings. This makes it well-suited to our goal of extracting visually recognizable patterns—such as QR codes—under the guise of standard interpretability outputs. Importantly, LIME itself is not our core contribution. Rather, our key idea is to leverage explanation tools as a channel for watermark extraction—embedding ownership signals in a way that blends seamlessly into typical explanation workflows, thereby enhancing stealthiness and reducing the risk of detection by adversaries. As shown in \autoref{fig:overview}, when masking is applied according to the black regions of the QR code, the feature attribution values in the corresponding unmasked regions become zero. To further improve robustness and interpretability during QR code extraction, we incorporate a regularization term $\Omega(g)$, which enforces non-negativity on the attribution weights:
\begin{equation}
    \Omega(g) = \sum_{i} \max (0,-W_{i}),
\end{equation}
Thus, the final optimization objective is given by:
\begin{equation}
    g^{*} = \arg \min_{g} L(f,g) + \Omega(g),
\end{equation}
While we adopt LIME for its simplicity and strong alignment with our design goals, other explanation methods—such as SHAP \cite{lundberg2017unified}, Integrated Gradients \cite{sundararajan2017axiomatic}, or Grad-CAM \cite{selvaraju2020grad} could also be used in principle, making our framework adaptable to various tasks and model types.
In practical applications, the optimized attribution matrix $W$ is normalized and binarized to produce a mask consisting of 0s and 1s, which can be directly used for ownership verification through QR code recognition.

\section{Experiment}
\label{sec:experiment}

\subsection{Experimental Settings} 
\subsubsection{Datasets}

We conduct experiments on four large-scale medical image segmentation datasets, covering cardiac, polyp, echocardiography, and fundus images, to ensure our proposed method can be widely applied across various medical segmentation scenarios. The detailed data splits for training, validation, and testing are summarized in Table~\ref{tab:dataset}. In these four datasets, 50\% of the images are randomly selected to embed triggers during training and testing.

\textbf{UKBB CMR Dataset:} 
The UK Biobank (UKBB)\footnote{This research has been conducted using the UK Biobank Resource under application number [54078].} is a large-scale, population-based biomedical database and research resource that contains detailed health information on approximately half a million participants from the United Kingdom. Our study utilized a subset of UKBB comprising four-chamber and two-chamber view CMR images at end-diastolic (ED) and end-systolic (ES) time points. Two-chamber view image with the left atrium (LA) endocardial region annotated. Four-chamber view image with the left atrium (LA) and right atrium (RA) endocardial regions annotated. The dataset contains a total of 18,160 images, of which 10,210 are used for training, 2,553 for validation, and 5,397 for testing.

\textbf{SEG Dataset:} SEG Dataset (Meng et al. \cite{meng2024multi} ) a total of 3,588 color fundus samples from six public datasets: \textit{Refuge} \cite{orlando2020refuge}, \textit{Drishti-GS} \cite{sivaswamy2014drishti}, \textit{ORIGA} \cite{zhang2010origa}, \textit{RIGA} \cite{almazroa2018retinal}, \textit{RIMONE} \cite{fumero2011rim}, \textit{G1020} \cite{bajwa2020g1020}. Those datasets provide the fundus images and the Optic Disc(OD) \& Optic Cup(OC) mask ground truths. A random selection of 715 fundus images was used as an external test dataset, with the remaining 2,870 images used for the five-fold training and cross-validation.

\textbf{EchoNet Dataset:} EchoNet \cite{ouyang2020video} is a large public dataset with 2D apical four-chamber echocardiography video sequences. Experienced cardiologists manually labelled a pair of two frames (end-systole and end-diastole) in each sequence. The annotations include the boundaries of the LVendo, LVepi, and LA at the ED and ES phases. Following the official dataset split, among the 10,030 pairs of images, 8,753 were used for five-fold training and cross-validation and 1,277 image pairs for testing.

\textbf{PraNet Dataset:} PraNet \cite{fan2020pranet} dataset is currently a widely used polyp segmentation colonoscopy imaging dataset. It contains five polyp segmentation datasets: \textit{ETIS} \cite{silva2014toward}, \textit{CVC-ClinicDB/CVC-612}  \cite{bernal2015wm}, \textit{CVC-ColonDB}  \cite{tajbakhsh2015automated}, \textit{EndoScene}  \cite{vazquez2017benchmark}, and \textit{Kvasir} \cite{jha2020kvasir}. The training set consists of 1,450 samples, including 900 images from \textit{Kvasir} and 550 images from \textit{CVC-ClinicDB/CVC-612}. The test set includes a total of 798 samples: the remaining 100 images from \textit{Kvasir}, 62 images from \textit{CVC-ClinicDB/CVC-612}, the complete \textit{CVC-ColonDB} dataset (380 samples), the complete \textit{ETIS} (196 samples), and 60 images from the \textit{EndoScene} (Also known as \textit{CVC-300}).

\begin{table}[h]
    \centering
    \caption{Summary of dataset splits used in our experiments. For each dataset, the numbers of samples in the training, validation, and testing sets are listed.}
    \tabcolsep=2mm
    \renewcommand{\arraystretch}{1.2}
    \begin{tabular}{l|c|c|c}
        \Xhline{2pt}  
        Dataset & Training Samples & Validation Samples & Test Samples\\
        \Xhline{1pt} 
        UKBB& 10,210 &2,553& 5,397 \\ 
        SEG & 2,296&574  & 715  \\  %total 1078 5:1 split
        EchoNet & 7,002  &1,751& 1277 \\ 
        PraNet& 1,160 &290& 798 \\ 
        \Xhline{2pt}  
    \end{tabular}
    \label{tab:dataset}
\end{table}
\subsubsection{Model}
We verify the effectiveness of our StealthMark method on popular segmentation models: \textbf{nnUNet} \cite{isensee2021nnu}, \textbf{Swin-UNet} \cite{cao2022swin}, \textbf{Trans-UNet} \cite{chen2021transunet}, \textbf{SAM} \cite{kirillov2023segment}, and \textbf{MedSAM} \cite{ma2024segment}.

\subsubsection{Implement Details}
Experiments are conducted on a system with 4 $\times$ NVIDIA GeForce RTX 4090 GPUs and an Intel Xeon Silver 4208 CPU, using the AdamW optimizer and a fixed learning rate of 0.0001. All models are trained for the same number of epochs. Except for MedSAM (1024 $\times$ 1024 image size, batch size 4, all models use 256 $\times$ 256 image size with a batch size of 32. We adopt four trigger types: Noise, Text, Patch, and Black Edge, with a default size of 1/64 of the image resolution. The ablation study further examines the effect of varying trigger sizes on performance. In the experiments presented in \autoref{tab:cvwm}, \autoref{eq4} is applied to constrain the background, with the hyperparameter $\delta$ set to 0.05 by default, while no constraint is imposed on the foreground. In the ablation study, we further explore the effect of $\delta$ in \autoref{eq4} and $\tau$ in \autoref{eq5} on model performance.

\subsubsection{Evaluation Metrics}
For the reported $p$-values, a chi-squared test of independence is used to verify that benign and triggered predictions originate from significantly different distributions. We evaluate performance from three perspectives: segmentation metrics (Dice, AUC, Volume Similarity), statistical test metrics, and adversarial attack metrics (ASR), details as follows,
\vspace{10pt}
\\
\textbf{Dice coefficient (Dice):} 
Measures the overlap between the predicted segmentation and the ground truth segmentation, normalized by their combined size. Here, $A$ represents the set of pixels (or voxels) in the predicted segmentation mask, and $B$ represents the set of pixels (or voxels) in the ground truth segmentation mask. Higher values indicate greater similarity.  
\begin{equation}
    \operatorname{Dice}(A, B) = \frac{2 \times |A \cap B|}{|A| + |B|}.
\end{equation}

\noindent \textbf{Area Under the Curve (AUC):} Measures the area under the Receiver Operating Characteristic (ROC) curve, which plots the true positive rate (TPR) against the false positive rate (FPR). Here, \(\text{TPR} = \frac{\text{TP}}{\text{TP} + \text{FN}}\) is the true positive rate, with \(\text{TP}\) as true positives and \(\text{FN}\) as false negatives, and \(\text{FPR} = \frac{\text{FP}}{\text{FP} + \text{TN}}\) is the false positive rate, with \(\text{FP}\) as false positives and \(\text{TN}\) as true negatives. A higher AUC indicates better discriminative performance. The AUC is computed as:
\begin{equation}
    \text{AUC} = \int_0^1 \text{TPR}(\text{FPR}^{-1}(x)) \, dx.
\end{equation}

\noindent\textbf{Volume Similarity (VS):} Measures the volumetric similarity between the predicted and ground truth segmentation regions by evaluating the difference in voxel counts. It is commonly used to assess the accuracy of tumor or organ volume estimation in medical imaging. Let $V_p$ denote the number of voxels in the predicted segmentation mask, and $V_g$ denote the number of voxels in the ground truth segmentation mask. The Volume Similarity is defined as:
\begin{equation}
    \mathrm{VS} = 1 - \frac{|V_p - V_g|}{V_p + V_g}.
\end{equation}

\noindent\textbf{Attack Success Rate (ASR):} The ratio of successful attack attempts to the total number of attempts, where $N_{\text{success}}$ is the number of successful attempts and $N_{\text{total}}$ is the total number of attempts. Higher ASR indicates a more effective attack.
\begin{equation}
    \operatorname{ASR} = \frac{N_{\text{success}}}{N_{\text{total}}}.
\end{equation}
% \noindent\textbf{Dice coefficient (Dice):} Dice is commonly used in image segmentation. A and B represent the sets of pixels belonging to the predicted segmentation result and the ground truth segmentation, respectively. Calculate the overlap between two sets relative to their combined size, with higher values indicating greater similarity.
% \begin{equation}
%     \small
%     \operatorname{Dice}(A, B)=\frac{2 \times|A \cap B|}{|A|+|B|}.
% \end{equation}
% \noindent \textbf{Intersection over Union (IoU):} IoU is another widely used metric in image segmentation. It measures the overlap between two sets divided by their union, with higher values indicating better agreement between predicted and ground truth regions.
% \begin{equation}
%     \small
%     \operatorname{IoU}(A, B)=\frac{|A \cap B|}{|A \cup B|}.
% \end{equation}
% \noindent \textbf{Attack Success Rate (ASR)} ASR is commonly used to evaluate the success rate of attacks in the field of computer security, especially in adversarial attacks, cryptography, or network security testing. It is defined as the ratio of successful attacks to the total number of attack attempts, where $N_{\text {success }}$ represents the number of successful attack attempts, and $N_{\text {total }}$ represents the total number of attack attempts.
% \begin{equation}
%     \small
%     \operatorname{ASR} = \frac{N_{\text{success}}}{N_{\text{total}}},
% \end{equation}
% Where a higher ASR indicates a more effective attack strategy.
\begin{table}
    \centering
    \tabcolsep=1.2mm
    \renewcommand{\arraystretch}{1.2}
    \caption{The segmentation scene attack evaluation of StealthMark on mainstream segmentation models}
    % \textcolor{red}{The column "ASR p-values" is all the same, the redundancy is too high, please reorganize this table}
    \label{tab:cvwm}
    \scalebox{0.85}{
    \begin{tabular}{cccccc}
    \Xhline{1pt}  
    \rowcolor{lightgray} 
    Model & Metric$\downarrow$ Trigger$\rightarrow$ & Noise & Text Patch & Color Patch & Black Edge\\
    \Xhline{1pt}
    \rowcolor{cyan!20}
    \multicolumn{6}{c}{\textbf{Dataset: UKBB}}\\
    \Xhline{1pt} 
        \multirow{2}{*}{nnUNet}
        &ASR  & 85.75 & 94.42 & 97.91& 98.11\\
        & p-value  & $10^{-13}$ & $10^{-13}$ & $10^{-13}$ & $10^{-13}$\\
        \cline{2-6}  
        \Xhline{1pt}
        \multirow{2}{*}{Swin-UNet} 
        &ASR  & 89.12 & 96.87 &100& 99.96\\
        & p-value  & $10^{-13}$ & $10^{-13}$& $10^{-13}$&$10^{-13}$\\
        \cline{2-6} 
        \Xhline{1pt}
        \multirow{2}{*}{Trans-UNet} 
        &ASR  & 69.04& 80.21 &94.57 & 99.11\\
        & p-value  &$10^{-2}$ & $10^{-13}$  & $10^{-13}$ & $10^{-13}$ \\
        \cline{2-6} 
        \Xhline{1pt}
        \multirow{2}{*}{SAM} 
        &ASR  & 98.93& 96.61& 98.65&99.80\\
        & p-value &  $10^{-13}$ & $10^{-13}$& $10^{-13}$&$10^{-13}$\\
        \cline{2-6} 
        \Xhline{1pt}
                \multirow{2}{*}{MedSAM}
        &ASR  & 99.52 & 97.05& 100 & 100\\
        & p-value  & $10^{-13}$& $10^{-13}$ & $10^{-13}$& $10^{-13}$\\
        \cline{2-6} 
        \Xhline{1pt}
    \rowcolor{cyan!20}

    \multicolumn{6}{c}{\textbf{Dataset: SEG}}\\
    
    \Xhline{1pt}  
        \multirow{2}{*}{nnUNet}

        &ASR  &90.80 &98.65&99.32 &94.28 \\
        & p-value  & $10^{-13}$ & $10^{-13}$ & $10^{-13}$ & $10^{-13}$\\
        \cline{2-6}  
        \Xhline{1pt}
        \multirow{2}{*}{Swin-UNet} 
        &ASR  & 92.17 & 97.20 & 98.32  & 98.88\\
        & p-value  & $10^{-13}$ & $10^{-13}$ & $10^{-13}$ & $10^{-13}$\\
        \cline{2-6}  
        \Xhline{1pt}
        \multirow{2}{*}{Trans-UNet} &ASR  & 72.73  &84.76 &78.04 & 96.92\\
        & p-value  & $10^{-2}$ & $10^{-13}$ & $10^{-1}$ & $10^{-13}$\\
        \cline{2-6}  
        \Xhline{1pt}
        \multirow{2}{*}{SAM} &ASR  & 99.30& 98.68 & 99.30 &86.01\\
        & p-value  & $10^{-13}$ & $10^{-13}$ & $10^{-13}$ &$10^{-13}$ \\
        \cline{2-6} 
        \Xhline{1pt}
                \multirow{2}{*}{MedSAM}&ASR  & 99.34 & 98.82 & 99.30 & 85.91\\
        & p-value  & $10^{-13}$ & $10^{-13}$ & $10^{-13}$ & $10^{-6}$\\
        \cline{2-6}  
        \Xhline{1pt}
    \rowcolor{cyan!20}

            \multicolumn{6}{c}{\textbf{Dataset: EchoNet}}\\

    \Xhline{1pt}  
        \multirow{2}{*}{nnUNet} &ASR  & 93.62 & 99.21 & 95.48& 91.44\\
        & p-value  & $10^{-13}$ & $10^{-13}$ & $10^{-13}$ & $10^{-13}$\\
        \cline{2-6}  
        \Xhline{1pt}
        \multirow{2}{*}{Swin-UNet} &ASR  & 94.82  &97.61& 97.49 & 98.38\\
        & p-value  & $10^{-13}$& $10^{-13}$ & $10^{-13}$& $10^{-13}$\\
        \cline{2-6}  
        \Xhline{1pt}
        \multirow{2}{*}{Trans-UNet} &ASR  & 79.60 & 100& 94.74 & 90.81\\
        & p-value  & $10^{-13}$ & $10^{-13}$ & $10^{-13}$ & $10^{-13}$\\
        \cline{2-6}  
        \Xhline{1pt}
        \multirow{2}{*}{SAM} &ASR  & 99.51 & 99.43 & 99.92 & 99.84\\
        & p-value  & $10^{-13}$ & $10^{-13}$ & $10^{-13}$ & $10^{-13}$\\
        \cline{2-6}  
        \Xhline{1pt}
                \multirow{2}{*}{MedSAM} &ASR  & 99.50 & 98.99 & 100 & 99.68\\
        & p-value  & $10^{-13}$ & $10^{-13}$ & $10^{-13}$ & $10^{-13}$\\
        \cline{2-6}  
        \Xhline{1pt}
    \rowcolor{cyan!20}
            \multicolumn{6}{c}{\textbf{Dataset: PraNet}}\\
    \Xhline{1pt} 
        \multirow{2}{*}{nnUNet} &ASR  & 72.01 & 90.12 & 88.78 & 61.34\\
        & p-value  & $10^{-2}$ & $10^{-13}$ & $10^{-13}$ & $10^{-13}$\\
        \cline{2-6}  
        \Xhline{1pt}
        \multirow{2}{*}{Swin-UNet} &ASR  & 52.75 &72.09 &82.37 &58.14\\
        & p-value  & $10^{-1}$ & $10^{-2}$ & $10^{-13}$ & $10^{-13}$\\
        \cline{2-6}  
        \Xhline{1pt}
        \multirow{2}{*}{Trans-UNet} &ASR  & 51.53 &49.82  & 61.20&51.77\\
        & p-value   & $10^{-1}$ & $10^{-1}$ & $10^{-13}$ & $10^{-1}$\\
        \cline{2-6}  
        \Xhline{1pt}
        \multirow{2}{*}{SAM} &ASR  & 99.14 & 99.76 &98.16  & 76.74\\
        & p-value  & $10^{-13}$ & $10^{-13}$ & $10^{-13}$ & $10^{-1}$\\
        \cline{2-6}  
        \Xhline{1pt}
                \multirow{2}{*}{MedSAM}&ASR  & 98.81 &99.74 & 99.14& 74.85\\
        & p-value   & $10^{-13}$ & $10^{-13}$ & $10^{-13}$ & $10^{-13}$\\
        \cline{2-6}  
        \Xhline{1pt}
    \Xhline{1pt}  
    \end{tabular}
    }
\end{table}
\\

% First table {Clean, Patch, Text}
\begin{table*}[t!]
    \centering
    \tabcolsep=1.2mm
    \renewcommand{\arraystretch}{1.4}
    \caption{The comparison experiment between StealthMark and backdoor-based models watermark method under different triggers (Part I)}
    \resizebox{0.9\textwidth}{!}{
    \begin{tabular}{|l|c|ccc|c|ccc|ccc|}
        \Xhline{2pt}
        \multirow{2}{*}{Dataset} & \multirow{2}{*}{Model}
            & \multicolumn{3}{c|}{Clean Model}
            & Trigger$\rightarrow$
            & \multicolumn{3}{c|}{Patch \cite{zhang2018protecting}}
            & \multicolumn{3}{c|}{Text \cite{zhang2018protecting}} \\
        \cline{3-12}
        & & Dice & AUC & VS & Method$\downarrow$
        & Dice & AUC & VS
        & Dice & AUC & VS \\
        \hline
        % UKBB
        \multirow{10}{*}{UKBB}
            & \multirow{2}{*}{nnUNet} & \multirow{2}{*}{94.73} & \multirow{2}{*}{97.35} & \multirow{2}{*}{95.42} & Backdoor
                & 90.94\textcolor{cred}{\footnotesize (-4\%)} & 97.30\textcolor{cred}{\footnotesize (-0\%)} & 92.55\textcolor{cred}{\footnotesize (-3\%)}
                & 89.99\textcolor{cred}{\footnotesize (-5\%)} & 97.31\textcolor{cred}{\footnotesize (-0\%)} & 93.72\textcolor{cred}{\footnotesize (-2\%)}
                \\
            &&&&& Ours
                &94.16\textcolor{cgreen}{\footnotesize (-0.6\%)} & 97.30\textcolor{cgreen}{\footnotesize (-0\%)} & 93.51\textcolor{cgreen}{\footnotesize (-2\%)}
                & 93.87\textcolor{cgreen}{\footnotesize (-1\%)} & 97.34\textcolor{cgreen}{\footnotesize (-0\%)} & 93.62\textcolor{cgreen}{\footnotesize (-2\%)}
                \\
            \cline{3-12}
            & \multirow{2}{*}{Swin-UNet} & \multirow{2}{*}{93.30} & \multirow{2}{*}{99.78} & \multirow{2}{*}{91.42} & Backdoor
                & 84.91 \textcolor{cred}{\footnotesize (-9\%)} & 99.75\textcolor{cred}{\footnotesize (-0\%)} & 83.18\textcolor{cred}{\footnotesize (-9\%)}
                & 87.03 \textcolor{cred}{\footnotesize (-7\%)} & 99.73\textcolor{cred}{\footnotesize (-0\%)} & 85.07\textcolor{cred}{\footnotesize (-7\%)}
                \\
            &&&&& Ours
                & 93.30\textcolor{cgreen}{\footnotesize (-0.1\%)} & 99.78\textcolor{cgreen}{\footnotesize (-0\%)} & 91.42\textcolor{cgreen}{\footnotesize (-0\%)}
                & 91.92\textcolor{cgreen}{\footnotesize (-2\%)} & 99.77 \textcolor{cgreen}{\footnotesize (-0\%)} & 89.83\textcolor{cgreen}{\footnotesize (-1.7\%)}
                \\
            \cline{3-12}
            & \multirow{2}{*}{Trans-UNet} & \multirow{2}{*}{90.79} & \multirow{2}{*}{99.65} & \multirow{2}{*}{88.81} & Backdoor
                & 90.09 \textcolor{cred}{\footnotesize (-0\%)} & 99.66\textcolor{cred}{\footnotesize (-0\%)} & 88.76\textcolor{cgreen}{\footnotesize (-0\%)}
                & 84.81 \textcolor{cred}{\footnotesize (-7\%)} & 98.79\textcolor{cred}{\footnotesize (-0.9\%)} & 83.55\textcolor{cred}{\footnotesize (-6\%)}
                \\
            &&&&& Ours
                & 90.70 \textcolor{cgreen}{\footnotesize (-0\%)} & 99.69\textcolor{cgreen}{\footnotesize (-0\%)} & 83.79\textcolor{cred}{\footnotesize (-5\%)}
                & 87.31\textcolor{cgreen}{\footnotesize (-4\%)} & 99.51 \textcolor{cgreen}{\footnotesize (-0.1\%)} & 85.43\textcolor{cgreen}{\footnotesize (-4\%)}
                \\
            \cline{3-12}
            & \multirow{2}{*}{SAM} & \multirow{2}{*}{91.87} & \multirow{2}{*}{99.73} & \multirow{2}{*}{90.80} & Backdoor
                & 92.28 \textcolor{cred}{\footnotesize (-3\%)} & 99.72\textcolor{cred}{\footnotesize (-0\%)} & 90.50\textcolor{cgreen}{\footnotesize (-0\%)}
                & 93.07 \textcolor{cred}{\footnotesize (-1\%)} & 99.72\textcolor{cred}{\footnotesize (-0\%)} & 90.62\textcolor{cred}{\footnotesize (-0\%)}
                \\
            &&&&& Ours
                & 91.81 \textcolor{cgreen}{\footnotesize (-1\%)} & 99.73\textcolor{cgreen}{\footnotesize (-0\%)} & 89.73\textcolor{cred}{\footnotesize (-1\%)}
                & 92.66\textcolor{cgreen}{\footnotesize (-0.5\%)} & 99.73\textcolor{cgreen}{\footnotesize (-0\%)} & 91.24\textcolor{cgreen}{\footnotesize (-0\%)}
                \\
            \cline{3-12}
            & \multirow{2}{*}{MedSAM} & \multirow{2}{*}{94.79} & \multirow{2}{*}{99.78} & \multirow{2}{*}{93.45} & Backdoor
                & 92.39 \textcolor{cred}{\footnotesize (-3\%)} & 99.75\textcolor{cred}{\footnotesize (-0\%)} & 93.20\textcolor{cred}{\footnotesize (-0\%)}
                & 93.84\textcolor{cred}{\footnotesize (-1\%)} & 99.72\textcolor{cred}{\footnotesize (-0\%)} & 93.41\textcolor{cred}{\footnotesize (-0\%)}
                \\
            &&&&& Ours
                & 94.40\textcolor{cgreen}{\footnotesize (-0.3\%)} & 99.77\textcolor{cgreen}{\footnotesize (-0\%)} & 93.40\textcolor{cgreen}{\footnotesize (-0\%)}
                & 94.69\textcolor{cgreen}{\footnotesize (-0\%)} & 99.73\textcolor{cgreen}{\footnotesize (-0\%)} & 93.40\textcolor{cgreen}{\footnotesize (-0\%)}
                \\
            \cline{3-12}
            \hline
        % SEG
        \multirow{10}{*}{SEG}
            & \multirow{2}{*}{nnUNet} & \multirow{2}{*}{50.10} & \multirow{2}{*}{75.40} & \multirow{2}{*}{93.50} & Backdoor
                & 48.21\textcolor{cred}{\footnotesize (-4\%)} & 74.64\textcolor{cred}{\footnotesize (-1\%)} & 90.48\textcolor{cred}{\footnotesize (-3\%)}
                & 49.09 \textcolor{cred}{\footnotesize (-2\%)} & 73.89\textcolor{cred}{\footnotesize (-2\%)} & 93.02\textcolor{cred}{\footnotesize (-0.5\%)}
                \\
            &&&&& Ours
                & 50.10\textcolor{cgreen}{\footnotesize (-0\%)} & 75.40 \textcolor{cgreen}{\footnotesize (-0\%)} & 93.44\textcolor{cgreen}{\footnotesize (-0\%)}
                & 49.75 \textcolor{cgreen}{\footnotesize (-0.7\%)} & 75.39 \textcolor{cgreen}{\footnotesize (-0\%)} & 93.52\textcolor{cgreen}{\footnotesize (-0\%)}
                \\
            \cline{3-12}
            & \multirow{2}{*}{Swin-UNet} & \multirow{2}{*}{69.34} & \multirow{2}{*}{95.37} & \multirow{2}{*}{75.69} & Backdoor
                & 65.90\textcolor{cred}{\footnotesize (-5\%)} & 93.79\textcolor{cred}{\footnotesize (-1.6\%)} & 75.61\textcolor{cred}{\footnotesize (-0\%)}
                & 64.47\textcolor{cred}{\footnotesize (-7\%)} & 93.89\textcolor{cred}{\footnotesize (-1.5\%)} & 74.59\textcolor{cgreen}{\footnotesize (-1.4\%)}
                \\
            &&&&& Ours
                & 68.91 \textcolor{cgreen}{\footnotesize (-0.7\%)} & 95.05\textcolor{cgreen}{\footnotesize (-0.3\%)} & 75.61\textcolor{cgreen}{\footnotesize (-0\%)}
                & 66.06\textcolor{cgreen}{\footnotesize (-5\%)} & 95.03\textcolor{cgreen}{\footnotesize (-0.3\%)} & 73.12\textcolor{cred}{\footnotesize (-3\%)}
                \\
            \cline{3-12}
            & \multirow{2}{*}{Trans-UNet} & \multirow{2}{*}{55.55} & \multirow{2}{*}{90.73} & \multirow{2}{*}{58.72} & Backdoor
                & 47.90 \textcolor{cred}{\footnotesize (-14\%)} & 80.08\textcolor{cred}{\footnotesize (-11\%)} & 48.10\textcolor{cred}{\footnotesize (-18\%)}
                & 51.94 \textcolor{cred}{\footnotesize (-7\%)} & 87.45\textcolor{cred}{\footnotesize (-1.4\%)} & 55.93\textcolor{cred}{\footnotesize (-4\%)}
                \\
            &&&&& Ours
                & 55.41\textcolor{cgreen}{\footnotesize (-0.1\%)} & 90.53 \textcolor{cgreen}{\footnotesize (-0.2\%)} & 58.23\textcolor{cgreen}{\footnotesize (-1\%)}
                & 54.14\textcolor{cgreen}{\footnotesize (-1.3\%)} & 89.56 \textcolor{cgreen}{\footnotesize (-0\%)} & 57.68\textcolor{cgreen}{\footnotesize (-2\%)}
                \\
            \cline{3-12}
            & \multirow{2}{*}{SAM} & \multirow{2}{*}{60.73} & \multirow{2}{*}{94.17} & \multirow{2}{*}{69.23} & Backdoor
                & 48.31 \textcolor{cred}{\footnotesize (-21\%)} & 72.02\textcolor{cred}{\footnotesize (-23\%)} & 48.31\textcolor{cred}{\footnotesize (-30\%)}
                & 57.77 \textcolor{cred}{\footnotesize (-5\%)} & 93.23\textcolor{cred}{\footnotesize (-1\%)} & 65.44\textcolor{cred}{\footnotesize (-5\%)}
                \\
            &&&&& Ours
                & 60.33 \textcolor{cgreen}{\footnotesize (-0.8\%)} & 94.12\textcolor{cgreen}{\footnotesize (-0\%)} & 67.87\textcolor{cgreen}{\footnotesize (-2\%)}
                & 58.68 \textcolor{cgreen}{\footnotesize (-4\%)} & 93.82\textcolor{cgreen}{\footnotesize (-0.4\%)} & 65.98\textcolor{cgreen}{\footnotesize (-4.7\%)}
                \\
            \cline{3-12}
            & \multirow{2}{*}{MedSAM} & \multirow{2}{*}{67.36} & \multirow{2}{*}{96.72} & \multirow{2}{*}{77.85} & Backdoor
                & 53.42\textcolor{cred}{\footnotesize (-20\%)} &82.21\textcolor{cred}{\footnotesize (-15\%)} & 56.05\textcolor{cred}{\footnotesize (-28\%)}
                & 63.31\textcolor{cred}{\footnotesize (-6\%)} & 95.75\textcolor{cred}{\footnotesize (-1\%)} & 75.51\textcolor{cred}{\footnotesize (-3\%)}
                \\
            &&&&& Ours
                & 67.12\textcolor{cgreen}{\footnotesize (-0.3\%)} & 96.68\textcolor{cgreen}{\footnotesize (-0\%)} & 75.51\textcolor{cgreen}{\footnotesize (-3\%)}
                & 65.33\textcolor{cgreen}{\footnotesize (-3\%)} & 96.70\textcolor{cgreen}{\footnotesize (-0\%)} & 76.29\textcolor{cgreen}{\footnotesize (-2\%)}
                \\
            \cline{3-12}
            \hline
        % EchoNet
        \multirow{10}{*}{EchoNet}
            & \multirow{2}{*}{nnUNet} & \multirow{2}{*}{92.28} & \multirow{2}{*}{96.08} & \multirow{2}{*}{91.33} & Backdoor
                & 90.76\textcolor{cred}{\footnotesize (-2\%)} & 96.00\textcolor{cred}{\footnotesize (-0\%)} & 91.33\textcolor{cred}{\footnotesize (-1\%)}
                & 91.26\textcolor{cred}{\footnotesize (-1\%)} & 96.12\textcolor{cred}{\footnotesize (-0\%)} & 91.32\textcolor{cred}{\footnotesize (-0\%)}
                \\
            &&&&& Ours
                & 92.27\textcolor{cgreen}{\footnotesize (-0\%)} & 96.05\textcolor{cgreen}{\footnotesize (-0\%)} & 91.35\textcolor{cgreen}{\footnotesize (-0\%)}
                & 92.24\textcolor{cgreen}{\footnotesize (-0\%)} & 96.08\textcolor{cgreen}{\footnotesize (-0\%)} & 91.30\textcolor{cgreen}{\footnotesize (-0\%)}
                \\
            \cline{3-12}
            & \multirow{2}{*}{Swin-UNet} & \multirow{2}{*}{95.37} & \multirow{2}{*}{99.82} & \multirow{2}{*}{94.60} & Backdoor
                & 95.16 \textcolor{cred}{\footnotesize (-2\%)} & 99.76\textcolor{cred}{\footnotesize (-0\%)} & 94.37\textcolor{cred}{\footnotesize (-0\%)}
                & 95.38 \textcolor{cred}{\footnotesize (-3\%)} & 99.80\textcolor{cred}{\footnotesize (-0\%)} & 94.60\textcolor{cred}{\footnotesize (-0\%)}
                \\
            &&&&& Ours
                & 95.40\textcolor{cgreen}{\footnotesize (-0\%)} & 99.80\textcolor{cgreen}{\footnotesize (-0\%)} & 94.63\textcolor{cgreen}{\footnotesize (-0\%)}
                & 95.33 \textcolor{cgreen}{\footnotesize (-0\%)} & 99.80\textcolor{cgreen}{\footnotesize (-0\%)} & 94.60\textcolor{cgreen}{\footnotesize (-0\%)}
                \\
            \cline{3-12}
            & \multirow{2}{*}{Trans-UNet} & \multirow{2}{*}{94.27} & \multirow{2}{*}{99.73} & \multirow{2}{*}{93.50} & Backdoor
                & 94.04 \textcolor{cred}{\footnotesize (-2\%)} & 99.68\textcolor{cred}{\footnotesize (-0\%)} & 93.19\textcolor{cred}{\footnotesize (-3\%)}
                & 93.40 \textcolor{cred}{\footnotesize (-3\%)} & 99.55 \textcolor{cred}{\footnotesize (-0.2\%)} & 92.60\textcolor{cred}{\footnotesize (-1\%)}
                \\
            &&&&& Ours
                & 94.59 \textcolor{cgreen}{\footnotesize (-0\%)} & 99.73\textcolor{cgreen}{\footnotesize (-0\%)} & 93.80\textcolor{cgreen}{\footnotesize (-0\%)}
                & 94.15\textcolor{cgreen}{\footnotesize (-0\%)} & 99.67 \textcolor{cgreen}{\footnotesize (-0\%)} & 93.36\textcolor{cgreen}{\footnotesize (-0.1\%)}
                \\
            \cline{3-12}
            & \multirow{2}{*}{SAM} & \multirow{2}{*}{95.30} & \multirow{2}{*}{99.79} & \multirow{2}{*}{94.53} & Backdoor
                & 72.26 \textcolor{cred}{\footnotesize (-14\%)} & 38.53\textcolor{cred}{\footnotesize (-61\%)} & 72.26\textcolor{cred}{\footnotesize (-24\%)}
                & 95.21 \textcolor{cred}{\footnotesize (-1\%)} & 99.79 \textcolor{cred}{\footnotesize (-0.4\%)} & 94.41\textcolor{cred}{\footnotesize (-0\%)}
                \\
            &&&&& Ours
                & 95.39 \textcolor{cgreen}{\footnotesize (-0\%)} & 99.79\textcolor{cgreen}{\footnotesize (-0\%)} & 94.66\textcolor{cgreen}{\footnotesize (-0\%)}
                & 95.44\textcolor{cgreen}{\footnotesize (-0\%)} & 99.80 \textcolor{cgreen}{\footnotesize (-0\%)} & 94.69\textcolor{cgreen}{\footnotesize (-0\%)}
                \\
            \cline{3-12}
            & \multirow{2}{*}{MedSAM} & \multirow{2}{*}{96.22} & \multirow{2}{*}{99.89} & \multirow{2}{*}{96.42} & Backdoor
                & 92.48 \textcolor{cred}{\footnotesize (-4\%)} & 98.89 \textcolor{cred}{\footnotesize (-1\%)} & 91.60 \textcolor{cred}{\footnotesize (-5\%)}
                & 95.25 \textcolor{cred}{\footnotesize (-2\%)} &99.88 \textcolor{cred}{\footnotesize (-0\%)} &  96.42 \textcolor{cred}{\footnotesize (-0\%)}
                \\
            &&&&& Ours
                & 96.15\textcolor{cgreen}{\footnotesize (-0\%)} & 99.85\textcolor{cgreen}{\footnotesize (-0\%)} & 96.44\textcolor{cgreen}{\footnotesize (-0\%)}
                
                & 96.21\textcolor{cgreen}{\footnotesize (-0\%)} &99.90 \textcolor{cgreen}{\footnotesize (-0\%)} & 96.42\textcolor{cgreen}{\footnotesize (-0\%)}
                \\
            \cline{3-12}
            \hline
        % PraNet
        \multirow{10}{*}{PraNet}
            & \multirow{2}{*}{nnUNet} & \multirow{2}{*}{71.39} & \multirow{2}{*}{89.74} & \multirow{2}{*}{71.03} & Backdoor
                & 55.51 \textcolor{cred}{\footnotesize (-22\%)} &52.94 \textcolor{cred}{\footnotesize (-41\%)} & 39.77\textcolor{cred}{\footnotesize (-44\%)}
                
                & 42.12 \textcolor{cred}{\footnotesize (-41\%)} & 59.23\textcolor{cred}{\footnotesize (-34\%)} & 39.06\textcolor{cred}{\footnotesize (-45\%)}
                \\
            &&&&& Ours
                & 71.34 \textcolor{cgreen}{\footnotesize (-0\%)} & 88.84 \textcolor{cgreen}{\footnotesize (-1\%)} &67.48\textcolor{cgreen}{\footnotesize (-5\%)}
                
                & 69.96 \textcolor{cgreen}{\footnotesize (-2\%)} & 89.74\textcolor{cgreen}{\footnotesize (-0\%)} & 68.60\textcolor{cgreen}{\footnotesize (-2\%)}
                \\
            \cline{3-12}
            & \multirow{2}{*}{Swin-UNet} & \multirow{2}{*}{73.18} & \multirow{2}{*}{92.08} & \multirow{2}{*}{72.90} & Backdoor
                & 32.10 \textcolor{cred}{\footnotesize (-29\%)} & 48.72 \textcolor{cred}{\footnotesize (-47\%)} & 34.05\textcolor{cred}{\footnotesize (-53\%)}
                & 27.92 \textcolor{cred}{\footnotesize (-38\%)} & 50.62\textcolor{cred}{\footnotesize (-45\%)} & 28.36\textcolor{cred}{\footnotesize (-61\%)}
                \\
            &&&&& Ours
                & 70.28 \textcolor{cgreen}{\footnotesize (-0\%)} & 90.03 \textcolor{cgreen}{\footnotesize (-2\%)} & 69.89\textcolor{cgreen}{\footnotesize (-4\%)}
                & 74.12\textcolor{cgreen}{\footnotesize (-0\%)} & 89.79\textcolor{cgreen}{\footnotesize (-2\%)} & 72.92\textcolor{cgreen}{\footnotesize (-0\%)}
                \\
            \cline{3-12}
            & \multirow{2}{*}{Trans-UNet} & \multirow{2}{*}{57.38} & \multirow{2}{*}{83.89} & \multirow{2}{*}{58.13} & Backdoor
                & 53.49 \textcolor{cred}{\footnotesize (-6\%)} & 48.97\textcolor{cred}{\footnotesize (-41\%)} & 30.80\textcolor{cred}{\footnotesize (-47\%)}
                & 41.74 \textcolor{cred}{\footnotesize (-27\%)} & 50.56\textcolor{cred}{\footnotesize (-39\%)} & 30.60\textcolor{cred}{\footnotesize (-47\%)}
                \\
            &&&&& Ours
                & 57.12 \textcolor{cgreen}{\footnotesize (-1\%)} & 82.61 \textcolor{cgreen}{\footnotesize (-1\%)} & 52.70\textcolor{cgreen}{\footnotesize (-9.3\%)}
                & 53.86 \textcolor{cgreen}{\footnotesize (-7\%)} & 80.60 \textcolor{cgreen}{\footnotesize (-4\%)} & 55.66\textcolor{cgreen}{\footnotesize (-4\%)}
                \\
            \cline{3-12}
            & \multirow{2}{*}{SAM} & \multirow{2}{*}{90.52} & \multirow{2}{*}{94.74} & \multirow{2}{*}{85.44} & Backdoor
                & 55.07 \textcolor{cred}{\footnotesize (-39\%)} & 51.76 \textcolor{cred}{\footnotesize (-45\%)} & 47.25\textcolor{cred}{\footnotesize (-44\%)}
                & 88.32 \textcolor{cred}{\footnotesize (-2\%)} & 50.94 \textcolor{cred}{\footnotesize (-46\%)}
                & 48.74\textcolor{cred}{\footnotesize (-42\%)}
                \\
            &&&&& Ours
                & 90.83\textcolor{cgreen}{\footnotesize (-0\%)} & 94.90\textcolor{cgreen}{\footnotesize (-0\%)} & 85.24\textcolor{cgreen}{\footnotesize (-0.2\%)}
                & 90.00 \textcolor{cgreen}{\footnotesize (-1\%)} & 94.99\textcolor{cgreen}{\footnotesize (-0\%)} & 85.43\textcolor{cgreen}{\footnotesize (-0\%)}
                \\
            \cline{3-12}
            & \multirow{2}{*}{MedSAM} & \multirow{2}{*}{91.94} & \multirow{2}{*}{95.65} & \multirow{2}{*}{88.10} & Backdoor
                & 58.95 \textcolor{cred}{\footnotesize (-36\%)} & 57.39\textcolor{cred}{\footnotesize (-40\%)} & 46.69\textcolor{cred}{\footnotesize (-47\%)}
                & 90.10 \textcolor{cred}{\footnotesize (-2\%)} & 84.17\textcolor{cred}{\footnotesize (-12\%)} & 68.71\textcolor{cred}{\footnotesize (-22\%)}
                \\
            &&&&& Ours
                & 90.86\textcolor{cgreen}{\footnotesize (-1\%)} &94.69 \textcolor{cgreen}{\footnotesize (-1\%)} &87.21\textcolor{cgreen}{\footnotesize (-1\%)}
                
                & 91.02 \textcolor{cgreen}{\footnotesize (-1\%)} & 95.40\textcolor{cgreen}{\footnotesize (-0\%)} &88.12 \textcolor{cgreen}{\footnotesize (-0\%)}
                \\
            \cline{3-12}
            \hline
        \Xhline{2pt}
    \end{tabular}
    }
    \label{tab:harmless_part1}
\end{table*}

% Table 2 {Clean, Noise, Black Edge}
\begin{table*}[t!]
    \centering
    \tabcolsep=1.2mm
    \renewcommand{\arraystretch}{1.4}
    \caption{The comparison experiment between StealthMark and backdoor-based models watermark method under different triggers (Part II)}
    \resizebox{0.9\textwidth}{!}{
    \begin{tabular}{|l|c|ccc|c|ccc|ccc|}
        \Xhline{2pt}
        \multirow{2}{*}{Dataset} & \multirow{2}{*}{Model}
            & \multicolumn{3}{c|}{Clean Model}
            & Trigger$\rightarrow$
            & \multicolumn{3}{c|}{Noise \cite{lounici2021yes}}
            & \multicolumn{3}{c|}{Black Edge} \\
        \cline{3-12}
        & & Dice & AUC & VS & Method$\downarrow$
        & Dice & AUC & VS
        & Dice & AUC & VS \\
        \hline
        % UKBB
        \multirow{10}{*}{UKBB}
            & \multirow{2}{*}{nnUNet} & \multirow{2}{*}{94.73} & \multirow{2}{*}{97.35} & \multirow{2}{*}{95.42} & Backdoor
                & 90.93\textcolor{cred}{\footnotesize (-4\%)} & 97.35\textcolor{cred}{\footnotesize (-0\%)} & 95.22\textcolor{cred}{\footnotesize (-0\%)}
                & 89.09\textcolor{cred}{\footnotesize (-6\%)} & 97.22\textcolor{cred}{\footnotesize (-0\%)} & 92.42\textcolor{cred}{\footnotesize (-3\%)}
                \\
            &&&&& Ours
                & 94.67\textcolor{cgreen}{\footnotesize (-0\%)} & 97.35\textcolor{cgreen}{\footnotesize (-0\%)} & 95.42\textcolor{cgreen}{\footnotesize (-0\%)}
                & 92.83\textcolor{cgreen}{\footnotesize (-2\%)} & 97.33\textcolor{cgreen}{\footnotesize (-0\%)} & 93.24\textcolor{cgreen}{\footnotesize (-2\%)}
                \\
            \cline{3-12}
            & \multirow{2}{*}{Swin-UNet} & \multirow{2}{*}{93.30} & \multirow{2}{*}{99.78} & \multirow{2}{*}{91.42} & Backdoor
                & 90.90 \textcolor{cred}{\footnotesize (-3\%)} & 99.76\textcolor{cred}{\footnotesize (-0\%)} & 88.75\textcolor{cgreen}{\footnotesize (-3\%)}
                & 94.27 \textcolor{cred}{\footnotesize (-6\%)} & 99.72\textcolor{cred}{\footnotesize (-0\%)} & 90.21\textcolor{cred}{\footnotesize (-1\%)}
                \\
            &&&&& Ours
                & 89.73 \textcolor{cgreen}{\footnotesize (-2\%)} & 99.76 \textcolor{cgreen}{\footnotesize (-0\%)} & 87.62\textcolor{cred}{\footnotesize (-4\%)}
                & 87.35\textcolor{cgreen}{\footnotesize (-0.2\%)} & 99.78 \textcolor{cgreen}{\footnotesize (-0\%)} & 90.21 \textcolor{cgreen}{\footnotesize (-1\%)}
                \\
            \cline{3-12}
            & \multirow{2}{*}{Trans-UNet} & \multirow{2}{*}{90.79} & \multirow{2}{*}{99.65} & \multirow{2}{*}{88.81} & Backdoor
                & 90.86 \textcolor{cred}{\footnotesize (-0\%)} & 99.38\textcolor{cred}{\footnotesize (-0\%)} & 88.67\textcolor{cred}{\footnotesize (-0\%)}
                & 87.41 \textcolor{cred}{\footnotesize (-4\%)} & 99.32 \textcolor{cred}{\footnotesize (-0.3\%)} & 85.61\textcolor{cred}{\footnotesize (-3.6\%)}
                \\
            &&&&& Ours
                & 90.77 \textcolor{cgreen}{\footnotesize (-0\%)} & 99.65\textcolor{cgreen}{\footnotesize (-0\%)} & 88.795\textcolor{cgreen}{\footnotesize (-0\%)}
                & 85.28 \textcolor{cgreen}{\footnotesize (-2\%)} & 99.61\textcolor{cgreen}{\footnotesize (-0\%)} & 86.58\textcolor{cgreen}{\footnotesize (-2\%)}
                \\
            \cline{3-12}
            & \multirow{2}{*}{SAM} & \multirow{2}{*}{91.87} & \multirow{2}{*}{99.73} & \multirow{2}{*}{90.80} & Backdoor
                & 92.15\textcolor{cred}{\footnotesize (-2\%)} & 99.70 \textcolor{cred}{\footnotesize (-0\%)} & 90.47\textcolor{cred}{\footnotesize (-0\%)}
                & 90.71\textcolor{cred}{\footnotesize (-4\%)} & 99.66 \textcolor{cred}{\footnotesize (-0\%)} & 88.61 \textcolor{cred}{\footnotesize (-2.4\%)}
                \\
            &&&&& Ours
                & 92.17 \textcolor{cgreen}{\footnotesize (-2\%)} & 99.73\textcolor{cgreen}{\footnotesize (-0\%)} & 90.54\textcolor{cgreen}{\footnotesize (-0\%)}
                & 92.16 \textcolor{cgreen}{\footnotesize (-2\%)} & 99.67\textcolor{cgreen}{\footnotesize (-0\%)} & 90.47\textcolor{cgreen}{\footnotesize (-0\%)}
                \\
            \cline{3-12}
            & \multirow{2}{*}{MedSAM} & \multirow{2}{*}{94.79} & \multirow{2}{*}{99.78} & \multirow{2}{*}{93.45} & Backdoor
                & 92.89 \textcolor{cred}{\footnotesize (-2\%)} & 99.75\textcolor{cred}{\footnotesize (-0\%)} & 93.41\textcolor{cgreen}{\footnotesize (-0\%)}
                & 92.98 \textcolor{cred}{\footnotesize (-2\%)} & 99.75\textcolor{cred}{\footnotesize (-0\%)} & 91.15 \textcolor{cred}{\footnotesize (-2\%)}
                \\
            &&&&& Ours
                & 93.84\textcolor{cgreen}{\footnotesize (-1\%)} & 99.80\textcolor{cgreen}{\footnotesize (-0\%)} & 93.20\textcolor{cred}{\footnotesize (-0\%)}
                & 93.93 \textcolor{cgreen}{\footnotesize (-1\%)} & 99.78\textcolor{cgreen}{\footnotesize (-0\%)} & 93.42 \textcolor{cgreen}{\footnotesize (-0\%)}
                \\
            \cline{3-12}
            \hline
        % SEG
        \multirow{10}{*}{SEG}
            & \multirow{2}{*}{nnUNet} & \multirow{2}{*}{50.10} & \multirow{2}{*}{75.40} & \multirow{2}{*}{93.50} & Backdoor
                & 49.14 \textcolor{cred}{\footnotesize (-2\%)} & 75.40\textcolor{cred}{\footnotesize (-0\%)} & 93.40\textcolor{cred}{\footnotesize (-0\%)}
                & 48.69\textcolor{cred}{\footnotesize (-3\%)} & 73.84\textcolor{cred}{\footnotesize (-2\%)} & 88.54\textcolor{cred}{\footnotesize (-5\%)}
                \\
            &&&&& Ours
                & 49.84 \textcolor{cgreen}{\footnotesize (-0.5\%)} & 75.42 \textcolor{cgreen}{\footnotesize (-0\%)} & 92.83\textcolor{cgreen}{\footnotesize (-0.7\%)}
                & 49.64\textcolor{cgreen}{\footnotesize (-1\%)} & 75.36 \textcolor{cgreen}{\footnotesize (-0\%)} & 91.05\textcolor{cgreen}{\footnotesize (-2.6\%)}
                \\
            \cline{3-12}
            & \multirow{2}{*}{Swin-UNet} & \multirow{2}{*}{69.34} & \multirow{2}{*}{95.37} & \multirow{2}{*}{75.69} & Backdoor
                & 64.42 \textcolor{cred}{\footnotesize (-7\%)} & 95.38\textcolor{cred}{\footnotesize (-0\%)} & 70.08\textcolor{cred}{\footnotesize (-7\%)}
                & 63.70 \textcolor{cred}{\footnotesize (-8\%)} & 93.69\textcolor{cred}{\footnotesize (-1.7\%)} & 72.68\textcolor{cred}{\footnotesize (-4\%)}
                \\
            &&&&& Ours
                & 68.09 \textcolor{cgreen}{\footnotesize (-2\%)} & 95.38 \textcolor{cgreen}{\footnotesize (-0\%)} & 71.16\textcolor{cgreen}{\footnotesize (-6\%)}
                & 68.81 \textcolor{cgreen}{\footnotesize (-1\%)} & 95.45 \textcolor{cgreen}{\footnotesize (-0\%)} & 75.82\textcolor{cgreen}{\footnotesize (-0\%)}
                \\
            \cline{3-12}
            & \multirow{2}{*}{Trans-UNet} & \multirow{2}{*}{55.55} & \multirow{2}{*}{90.73} & \multirow{2}{*}{58.72} & Backdoor
                & 54.31 \textcolor{cred}{\footnotesize (-2\%)} & 90.18 \textcolor{cred}{\footnotesize (-0\%)} & 58.57\textcolor{cred}{\footnotesize (-0\%)}
                & 47.96 \textcolor{cred}{\footnotesize (-16\%)} & 78.13 \textcolor{cred}{\footnotesize (-13\%)} & 47.96 \textcolor{cred}{\footnotesize (-18\%)}
                \\
            &&&&& Ours
                & 55.93\textcolor{cgreen}{\footnotesize (-0\%)} & 90.47 \textcolor{cgreen}{\footnotesize (-0\%)} & 57.66\textcolor{cgreen}{\footnotesize (-2\%)}
                & 56.36\textcolor{cgreen}{\footnotesize (-0\%)} & 90.22\textcolor{cgreen}{\footnotesize (-0\%)} & 58.35\textcolor{cgreen}{\footnotesize (-0.6\%)}
                \\
            \cline{3-12}
            & \multirow{2}{*}{SAM} & \multirow{2}{*}{60.73} & \multirow{2}{*}{94.17} & \multirow{2}{*}{69.23} & Backdoor
                & 48.41 \textcolor{cred}{\footnotesize (-26\%)} & 21.42\textcolor{cred}{\footnotesize (-77\%)} & 48.41\textcolor{cred}{\footnotesize (-30\%)}
                & 59.59 \textcolor{cred}{\footnotesize (-5\%)} & 94.09\textcolor{cred}{\footnotesize (-0\%)} & 67.10\textcolor{cred}{\footnotesize (-3\%)}
                \\
            &&&&& Ours
                & 58.41 \textcolor{cgreen}{\footnotesize (-5\%)} & 93.75 \textcolor{cgreen}{\footnotesize (-0.4\%)} & 65.23\textcolor{cgreen}{\footnotesize (-5.8\%)}
                & 59.44 \textcolor{cgreen}{\footnotesize (-2\%)} & 94.20 \textcolor{cgreen}{\footnotesize (-0\%)} & 66.69 \textcolor{cgreen}{\footnotesize (-3.6\%)}
                \\
            \cline{3-12}
            & \multirow{2}{*}{MedSAM} & \multirow{2}{*}{67.36} & \multirow{2}{*}{96.72} & \multirow{2}{*}{77.85} & Backdoor
                & 63.99\textcolor{cred}{\footnotesize (-5\%)} & 94.78\textcolor{cred}{\footnotesize (-2\%)} & 73.95\textcolor{cred}{\footnotesize (-5\%)}
                & 64.66\textcolor{cred}{\footnotesize (-4\%)} & 96.62\textcolor{cred}{\footnotesize (-0\%)} & 76.29\textcolor{cred}{\footnotesize (-2\%)}
                \\
            &&&&& Ours
                & 63.99 \textcolor{cgreen}{\footnotesize (-5\%)} & 95.75\textcolor{cgreen}{\footnotesize (-1\%)} & 74.34\textcolor{cgreen}{\footnotesize (-5\%)}
                & 66.68\textcolor{cgreen}{\footnotesize (-1\%)} & 96.58\textcolor{cgreen}{\footnotesize (-0\%)} & 76.44\textcolor{cgreen}{\footnotesize (-2\%)}
                \\
            \cline{3-12}
            \hline
        % EchoNet
        \multirow{10}{*}{EchoNet}
            & \multirow{2}{*}{nnUNet} & \multirow{2}{*}{92.28} & \multirow{2}{*}{96.08} & \multirow{2}{*}{91.33} & Backdoor
                & 91.44\textcolor{cred}{\footnotesize (-1\%)} & 96.00 \textcolor{cred}{\footnotesize (-0\%)} & 90.41\textcolor{cred}{\footnotesize (-1\%)}
                & 90.43 \textcolor{cred}{\footnotesize (-2\%)} & 95.59\textcolor{cred}{\footnotesize (-0.5\%)} & 90.69\textcolor{cgreen}{\footnotesize (-0.7\%)}
                \\
            &&&&& Ours
                & 92.28 \textcolor{cgreen}{\footnotesize (-0\%)} & 96.05\textcolor{cgreen}{\footnotesize (-0\%)} & 91.33\textcolor{cgreen}{\footnotesize (-0\%)}
                & 91.37\textcolor{cgreen}{\footnotesize (-1\%)} & 96.01\textcolor{cgreen}{\footnotesize (-0\%)} & 90.44 \textcolor{cred}{\footnotesize (-1\%)}
                \\
            \cline{3-12}
            & \multirow{2}{*}{Swin-UNet} & \multirow{2}{*}{95.37} & \multirow{2}{*}{99.82} & \multirow{2}{*}{94.60} & Backdoor
                & 95.38 \textcolor{cred}{\footnotesize (-1\%)} & 99.81 \textcolor{cred}{\footnotesize (-0\%)} & 94.60\textcolor{cred}{\footnotesize (-0\%)}
                & 94.85 \textcolor{cred}{\footnotesize (-2\%)} & 99.75\textcolor{cred}{\footnotesize (-0\%)} & 94.01\textcolor{cred}{\footnotesize (-0.6\%)}
                \\
            &&&&& Ours
                & 95.53 \textcolor{cgreen}{\footnotesize (-0\%)} & 99.79 \textcolor{cgreen}{\footnotesize (-0\%)} & 94.76\textcolor{cgreen}{\footnotesize (-0\%)}
                & 95.45 \textcolor{cgreen}{\footnotesize (-0\%)} & 99.81 \textcolor{cgreen}{\footnotesize (-0\%)} & 94.68 \textcolor{cgreen}{\footnotesize (-0\%)}
                \\
            \cline{3-12}
            & \multirow{2}{*}{Trans-UNet} & \multirow{2}{*}{94.27} & \multirow{2}{*}{99.73} & \multirow{2}{*}{93.50} & Backdoor
                & 93.27 \textcolor{cred}{\footnotesize (-2\%)} & 99.53\textcolor{cred}{\footnotesize (-0.2\%)} & 92.53\textcolor{cred}{\footnotesize (-1\%)}
                & 91.82 \textcolor{cred}{\footnotesize (-3\%)} & 99.32 \textcolor{cred}{\footnotesize (-0.4\%)} & 91.10 \textcolor{cred}{\footnotesize (-2.5\%)}
                \\
            &&&&& Ours
                & 94.17 \textcolor{cgreen}{\footnotesize (-0\%)} & 99.71 \textcolor{cgreen}{\footnotesize (-0\%)} & 93.39\textcolor{cgreen}{\footnotesize (-0.1\%)}
                & 94.17 \textcolor{cgreen}{\footnotesize (-0\%)} & 99.56 \textcolor{cgreen}{\footnotesize (-0.1\%)} & 93.39 \textcolor{cgreen}{\footnotesize (-0.1\%)}
                \\
            \cline{3-12}
            & \multirow{2}{*}{SAM} & \multirow{2}{*}{95.30} & \multirow{2}{*}{99.79} & \multirow{2}{*}{94.53} & Backdoor
                & 72.26 \textcolor{cred}{\footnotesize (-24\%)} & 39.47\textcolor{cred}{\footnotesize (-60\%)} & 72.26\textcolor{cred}{\footnotesize (-23\%)}
                & 95.29 \textcolor{cred}{\footnotesize (-1\%)} & 99.81 \textcolor{cred}{\footnotesize (-0.4\%)} & 94.54\textcolor{cred}{\footnotesize (-0\%)}
                \\
            &&&&& Ours
                & 95.45 \textcolor{cgreen}{\footnotesize (-0\%)} & 99.80\textcolor{cgreen}{\footnotesize (-0\%)} & 94.69\textcolor{cgreen}{\footnotesize (-0\%)}
                & 95.42 \textcolor{cgreen}{\footnotesize (-0\%)} & 99.79\textcolor{cgreen}{\footnotesize (-0\%)} & 94.66\textcolor{cgreen}{\footnotesize (-0\%)}
                \\
            \cline{3-12}
            & \multirow{2}{*}{MedSAM} & \multirow{2}{*}{96.22} & \multirow{2}{*}{99.89} & \multirow{2}{*}{96.42} & Backdoor
                & 92.74 \textcolor{cred}{\footnotesize (-4\%)} &99.89 \textcolor{cred}{\footnotesize (-1\%)} & 95.45 \textcolor{cred}{\footnotesize (-1\%)}
                & 95.16 \textcolor{cred}{\footnotesize (-2\%)} & 99.88\textcolor{cred}{\footnotesize (-0\%)} & 96.40\textcolor{cred}{\footnotesize (-0\%)}
                \\
            &&&&& Ours
                & 96.24\textcolor{cgreen}{\footnotesize (-0\%)} & 99.89\textcolor{cgreen}{\footnotesize (-0\%)} & 96.44\textcolor{cgreen}{\footnotesize (-0\%)}
                & 96.19 \textcolor{cgreen}{\footnotesize (-0\%)} & 99.88\textcolor{cgreen}{\footnotesize (-0\%)} & 96.45\textcolor{cgreen}{\footnotesize (-0\%)}
                \\
            \cline{3-12}
            \hline
        % PraNet
        \multirow{10}{*}{PraNet}
            & \multirow{2}{*}{nnUNet} & \multirow{2}{*}{71.39} & \multirow{2}{*}{89.74} & \multirow{2}{*}{71.03} & Backdoor
                & 56.39 \textcolor{cred}{\footnotesize (-21\%)} & 77.17\textcolor{cred}{\footnotesize (-14\%)} & 44.74\textcolor{cred}{\footnotesize (-37\%)}
                & 29.26 \textcolor{cred}{\footnotesize (-59\%)} & 49.35\textcolor{cred}{\footnotesize (-45\%)} & 34.09 \textcolor{cred}{\footnotesize (-52\%)}
                \\
            &&&&& Ours
                & 70.67\textcolor{cgreen}{\footnotesize (-1\%)} & 89.74\textcolor{cgreen}{\footnotesize (-0\%)} & 71.02\textcolor{cgreen}{\footnotesize (-0\%)}
                & 64.32\textcolor{cgreen}{\footnotesize (-10\%)} & 82.56\textcolor{cgreen}{\footnotesize (-8\%)} & 61.07\textcolor{cgreen}{\footnotesize (-14\%)}
                \\
            \cline{3-12}
            & \multirow{2}{*}{Swin-UNet} & \multirow{2}{*}{73.18} & \multirow{2}{*}{92.08} & \multirow{2}{*}{72.90} & Backdoor
                & 32.41 \textcolor{cred}{\footnotesize (-22\%)} & 48.83\textcolor{cred}{\footnotesize (-46\%)} & 34.92\textcolor{cred}{\footnotesize (-52\%)}
                & 19.06 \textcolor{cred}{\footnotesize (-58\%)} & 32.97\textcolor{cred}{\footnotesize (-64\%)} & 22.13\textcolor{cred}{\footnotesize (-69\%)}
                \\
            &&&&& Ours
                & 60.30 \textcolor{cgreen}{\footnotesize (-5\%)} & 85.75 \textcolor{cgreen}{\footnotesize (-6\%)} & 60.75\textcolor{cgreen}{\footnotesize (-16\%)}
                & 68.38 \textcolor{cgreen}{\footnotesize (-8\%)} & 85.57 \textcolor{cgreen}{\footnotesize (-7\%)} & 68.57\textcolor{cgreen}{\footnotesize (-6\%)}
                \\
            \cline{3-12}
            & \multirow{2}{*}{Trans-UNet} & \multirow{2}{*}{57.38} & \multirow{2}{*}{83.89} & \multirow{2}{*}{58.13} & Backdoor
                & 30.71 \textcolor{cred}{\footnotesize (-47\%)} & 55.76 \textcolor{cred}{\footnotesize (-33\%)} & 11.91\textcolor{cred}{\footnotesize (-79\%)}
                & 33.97 \textcolor{cred}{\footnotesize (-41\%)} & 60.04\textcolor{cred}{\footnotesize (-28\%)} & 21.42\textcolor{cred}{\footnotesize (-63\%)}
                \\
            &&&&& Ours
                & 57.71 \textcolor{cgreen}{\footnotesize (-0\%)} & 81.17 \textcolor{cgreen}{\footnotesize (-3\%)} & 57.80\textcolor{cgreen}{\footnotesize (-0\%)}
                & 56.07 \textcolor{cgreen}{\footnotesize (-3\%)} & 80.70 \textcolor{cgreen}{\footnotesize (-4\%)} & 57.70\textcolor{cgreen}{\footnotesize (-0.7\%)}
                \\
            \cline{3-12}
            & \multirow{2}{*}{SAM} & \multirow{2}{*}{90.52} & \multirow{2}{*}{94.74} & \multirow{2}{*}{85.44} & Backdoor
                & 82.40 \textcolor{cred}{\footnotesize (-10\%)} & 49.10 \textcolor{cred}{\footnotesize (-48\%)} & 45.55\textcolor{cred}{\footnotesize (-46\%)}
                & 65.71 \textcolor{cred}{\footnotesize (-27\%)} & 48.61 \textcolor{cred}{\footnotesize (-48\%)}
                & 44.31\textcolor{cred}{\footnotesize (-48\%)}
                \\
            &&&&& Ours
                & 90.14 \textcolor{cgreen}{\footnotesize (-0\%)} & 94.33 \textcolor{cgreen}{\footnotesize (-0\%)} & 85.61\textcolor{cgreen}{\footnotesize (-0\%)}
                & 90.02 \textcolor{cgreen}{\footnotesize (-1\%)} & 94.36 \textcolor{cgreen}{\footnotesize (-0.4\%)} & 85.44 \textcolor{cgreen}{\footnotesize (-0\%)}
                \\
            \cline{3-12}
            & \multirow{2}{*}{MedSAM} & \multirow{2}{*}{91.94} & \multirow{2}{*}{95.65} & \multirow{2}{*}{88.10} & Backdoor
                & 84.58 \textcolor{cred}{\footnotesize (-8\%)} & 72.69\textcolor{cred}{\footnotesize (-24\%)} & 71.36\textcolor{cred}{\footnotesize (-19\%)}
                & 69.87 \textcolor{cred}{\footnotesize (-24\%)} &60.26 \textcolor{cred}{\footnotesize (-37\%)} & 46.69 \textcolor{cred}{\footnotesize (-47\%)}
                \\
            &&&&& Ours
                & 91.85 \textcolor{cgreen}{\footnotesize (-0\%)} & 95.60\textcolor{cgreen}{\footnotesize (-0\%)} & 88.10 \textcolor{cgreen}{\footnotesize (-0\%)}
                
                & 90.10 \textcolor{cgreen}{\footnotesize (-2\%)} & 95.07\textcolor{cgreen}{\footnotesize (-0.6\%)} & 88.12\textcolor{cgreen}{\footnotesize (-0\%)}
                \\
            \cline{3-12}
            \hline
        \Xhline{2pt}
    \end{tabular}
    }
    \label{tab:harmless_part2}
\end{table*}

\subsection{Main Result}
\subsubsection{StealthMark Effectiveness Analysis}

StealthMark has demonstrated remarkable effectiveness and practicality in safeguarding the intellectual property of segmentation models, with its superior performance validated through attack evaluations across multiple datasets and models (see \autoref{tab:cvwm}). We tested four different triggers (Noise, Text, Patch, and Black Edge) on five mainstream segmentation models (nnUNet, Swin-UNet, Trans-UNet, SAM, and MedSAM) across four datasets (UKBB, SEG, EchoNet, and PraNet), achieving high attack success rates (ASRs). For instance, on the UKBB dataset, nnUNet achieved ASR of 85.75\%, 94.42\%, 97.91\%, and 98.11\% under Noise, Text Patch, Color Patch, and Black Edge attacks, respectively, while Swin-UNet and MedSAM approached 100\%. On the SEG and EchoNet datasets, most models achieved ASRs above 90\%, and SAM and MedSAM consistently exceeded 97\%, showcasing robust watermark resilience. Even on the PraNet dataset, where some models (e.g., Trans-UNet and Swin-UNet) exhibited relatively lower ASRs (49.82\% to 82.37\%), SAM and MedSAM still achieved very high ASR ranging from 98.16\% to 99.76\%. Extremely small p-values (\textit{e.g.} $10^{-13}$) further confirm the statistical significance of these findings. We also observed that large-scale pre-trained models, such as SAM \cite{kirillov2023segment}, exhibit strong robustness against implanted triggers across various datasets, yielding consistently high ASRs. In contrast, models that combine convolution and attention mechanisms, such as Trans-UNet, appear more susceptible to triggering interference. Moreover, on datasets like PraNet, where images inherently include patches, text, or black edges that resemble our triggers, the performance of our method was slightly lower than on other datasets. Overall, the high ASRs achieved by StealthMark ensure accurate model ownership verification, offering an efficient and reliable approach to protecting the intellectual property of segmentation models.

\subsubsection{StealthMark Harmlessness and Stealthiness Analysis}

We compare our method with existing backdoor-based model watermarking techniques, which are among the most representative and widely adopted strategies for verifying black-box model ownership. While these methods primarily differ in how their trigger sets are constructed, they share the core principle of embedding triggers that cause the model to produce specific, verifiable outputs. Our study follows established practices in the backdoor watermarking literature by evaluating our method under four representative trigger types: patch triggers, text overlays, random noise, and black edge patterns. Specifically, the patch and text overlay triggers are adapted from the method proposed by Zhang et al. \cite{zhang2018protecting}, while the random noise trigger follows the design introduced by Lounici et al. \cite{lounici2021yes}. The black edge pattern, widely used in recent studies \cite{shao2025explanation}, serves as a simple yet effective trigger type to simulate diverse real-world watermarking scenarios. Most backdoor-based watermarking methods are designed for image classification, where ownership is verified by forcing inputs with triggers to be misclassified into a target class. When directly applied to binary segmentation, these methods tend to produce extreme and unstable outputs: for example, background pixels that should be labeled as 0 are misclassified as 1, and foreground pixels that should be labeled as 1 are flipped to 0.

In contrast, our approach is more subtle, involving only a small adjustment to the background region’s minimum intensity (increased from 0 to 0.02) or the foreground region’s maximum intensity (decreased from 1 to 0.85). As shown in \autoref{tab:harmless_part1} and  \autoref{tab:harmless_part2}, experimental results demonstrate that our method achieves superior harmlessness. It is worth noting that datasets such as EchoNet and UKBB are relatively simple in terms of structure and content, which makes the models less sensitive to injected triggers—hence, backdoor-based watermarking methods tend to cause only minor performance degradation on these datasets. In contrast, datasets like PraNet and SEG present significantly more challenging segmentation tasks. Not only do they feature complex anatomical or pathological structures, but they also naturally contain patterns visually similar to certain types of triggers. This makes the models more vulnerable to overfitting on trigger patterns, leading to more severe performance degradation when traditional backdoor watermarking is applied. On datasets such as UKBB, SEG, EchoNet, and PraNet, using models like nnUNet, Swin-UNet, and MedSAM, the Dice, AUC and Volume Similarity scores under watermarking remain nearly identical to their clean counterparts. For instance, on the PraNet dataset with patch triggers, SAM’s Dice score decreases by only 1\%, Trans-UNet's Dice remains unchanged, and its AUC remains virtually unchanged. In contrast, backdoor-based watermarking methods can degrade segmentation performance by up to 39\%, severely limiting their practical deployment.

\begin{figure}[!ht]
\centering
\includegraphics[width=0.95\linewidth]{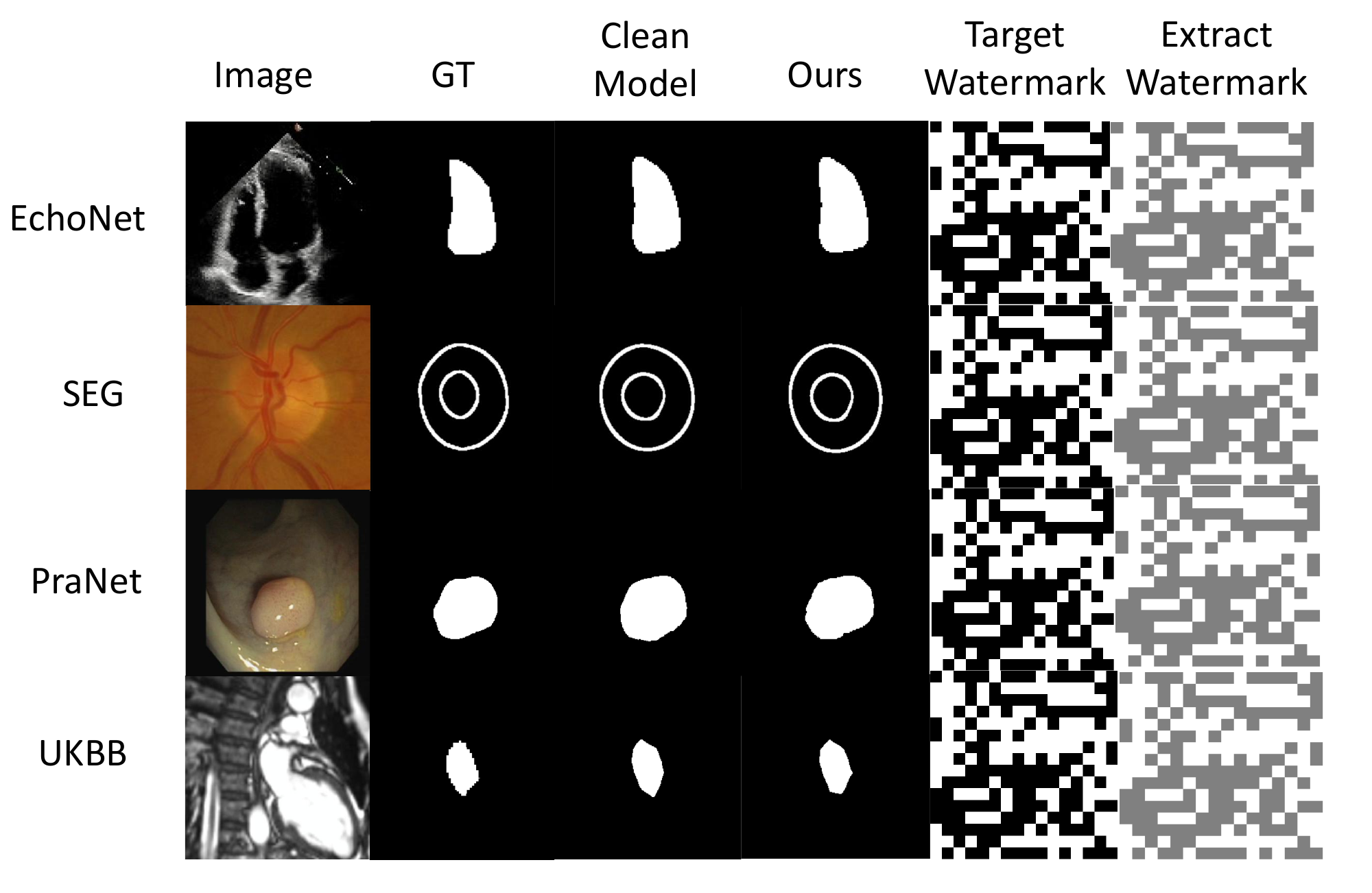}
\caption{Qualitative results on four medical segmentation datasets (EchoNet, SEG \cite{meng2024multi}, PraNet \cite{fan2020pranet}, and UKBB) using nnUNet \cite{isensee2021nnu}. For each dataset, we show the input image, ground truth mask (GT), prediction from the clean (unwatermarked) model, and prediction from our watermarked model. The right columns show the target watermark and the extracted watermark, confirming reliable ownership verification.}
\label{fig:V}
\end{figure}

\begin{figure*}
    \centering
    \includegraphics[width=0.9\linewidth]{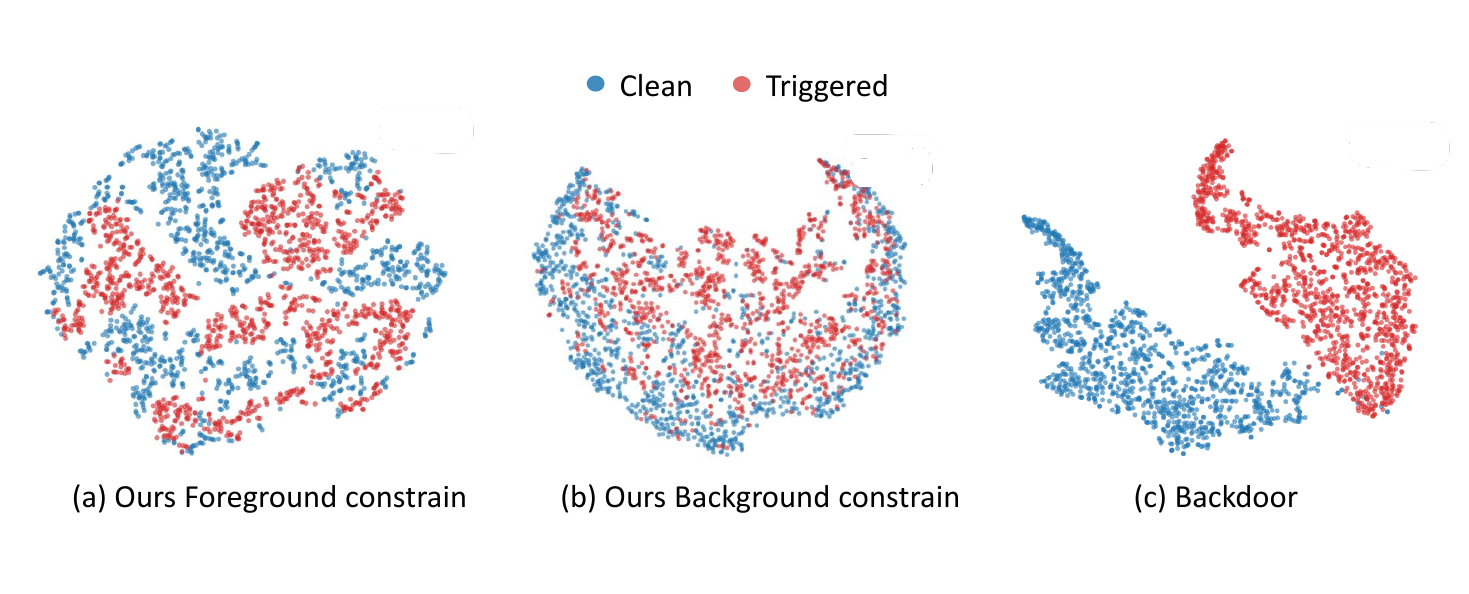}
    \caption{The t-SNE visualization of features (before the segmentation head) on the EchoNet test set using the SAM \cite{kirillov2023segment} model under patch-based trigger attacks. Subfigures \textbf{(a)} and \textbf{(b)} show examples of our method with foreground constraint and background constraint, respectively. In contrast, \textbf{(c)} shows a representative result from traditional backdoor methods (\textit{e.g.,} BadNet \cite{gu2019badnets}). Compared to the prior approach, our method yields significantly more entangled representations between clean and triggered samples, demonstrating higher stealthiness in the feature space.}
    \label{fig:tsne}
\end{figure*}

To provide an intuitive understanding of StealthMark’s harmlessness and reliability, we visualize segmentation results across four medical datasets (EchoNet \cite{ouyang2020video}, SEG \cite{meng2024multi}, PraNet \cite{fan2020pranet}, and UKBB) using nnUNet \cite{isensee2021nnu}, as shown in \autoref{fig:V}. Our method maintains almost identical segmentation quality compared to the clean model across all four datasets, confirming the harmlessness property of our watermark design. The extracted watermarks are visually consistent with their corresponding targets, demonstrating reliable and stable ownership verification.
Furthermore, to evaluate the feature-level stealthiness, we visualize the feature maps before the segmentation head in SAM-based models on the EchoNet dataset with patch trigger and show examples of our method with foreground constraint and background constraint, compared with the backdoor-based method (\autoref{fig:tsne}). Unlike traditional backdoor-based methods, our method maintains a high similarity between the clean and triggered feature distributions, making it more stealthy against detection or removal attacks.

\subsection{Ablation Study}
\label{sec:Abltion}

In this ablation study, we evaluate the effectiveness and stealthiness of our method under different parameter settings, focusing on trigger size and training parameters. Using the SAM model on the PraNet and EchoNet datasets, we systematically vary trigger size and batch size to analyze model performance and sensitivity to these configurations.

\subsubsection{Selecting the Optimal Patch Size for Effectiveness and Stealthiness}
\label{sec: abl_pacth}

We will determine the optimal Patch Size value of the model by ensuring the \textbf{Dice}, \textbf{AUC}, \textbf{Volume Similarity} and \textbf{ASR} of the segmentation performance indicators of the model after using our method. Our goal is to determine the optimal patch size that ensures guaranteed ASR performance while minimizing the impact on model segmentation performance. The experiment is conducted on PraNet with SAM as a case study, using an image resolution of 256 $\times$ 256. 

The results are presented in \autoref{tab:results_abl_1}. It can be seen from the experimental results that when the patch size is 4, the ASR of the model reaches 99.02\%, and the gap between Dice, AUC and Volume Similarity also tends to be within the normal range of the clean model results. From this, we determined the best patch size for StealthMark.

\begin{table}[htbp]
\renewcommand{\arraystretch}{0.9}
\centering
\caption{Results With Different trigger size (image size is 256) on PraNet Dataset}
\label{tab:results_abl_1}
\begin{tabular}{cccccc}
\toprule
Patch Size & Dice & AUC &VS &ASR \\
\midrule
2   & 90.27  & 95.05 &85.29& 53.98\\
\rowcolor{gray!15}
4 &90.28
     & 94.65  &85.84
     & 99.02  \\
8 & 90.09  
     & 94.69  & 85.40
     & 98.90  \\
16 & 90.42 
     & 94.69 &85.44
     & 99.51 \\
32 & 90.26
     & 93.68&85.57
     & 97.67 \\
\textit{Clean model} & 90.52 & 94.74 &85.44&-\\
\bottomrule
\end{tabular}
\end{table}

\subsubsection{Selecting the Optimal Hyper-Parameter \texorpdfstring{$\delta$}{delta} or \texorpdfstring{$\tau$}{tau} for Effectiveness and Stealthiness}
\label{sec: abl_para_m_n}

In this experiment, we conducted a case study using the SAM model's performance on the EchoNet dataset, employing a text trigger. We evaluated the effectiveness and stealthiness of our method by varying the parameters $\delta$ and $\tau$. The experimental results are presented in \autoref{tab:results_abl_2}. We initially defined the experimental range for these two hyperparameters as ($\tau$: 0.8–0.99, $\delta$: 0.001–0.1). When $\delta$ was set to 0.05, our method achieved an ASR of 99.60\% on EchoNet, while the segmentation performance remained indistinguishable from the clean model (Dice: 95.30, AUC: 99.80, VS: 93.59), ensuring effectiveness. Furthermore, the slight difference between 0.05 and the original value of 0 demonstrates strong stealthiness. Similarly, when $\tau$ was set to 0.85, the method achieved an ASR of 99.50\% on EchoNet, again without significantly affecting segmentation performance (Dice: 95.30, AUC: 99.85, VS: 94.56). These findings confirm that the chosen hyperparameters ($\delta=0.05$, $\tau=0.85$) provide a favorable trade-off between watermarking effectiveness and concealment. The high ASR, coupled with a negligible effect on segmentation quality, suggests that StealthMark can embed ownership without compromising the utility of the model or revealing its presence, thus fulfilling the core design goal of the method.

\begin{table}[!ht]
\renewcommand{\arraystretch}{0.9}
\centering
\caption{Results With Different Hyper-Parameter $\delta$ and $\tau$ on EchoNet Dataset}
\label{tab:results_abl_2}
\begin{tabular}{c c c c cc}
\toprule
$\delta$ & Dice &AUC&VS& ASR \\
\midrule
0.100    & 95.39&99.79&93.80 & 99.51\\
\rowcolor{gray!15}
0.050 
     & 95.30   &99.80&93.59
     & 99.50  \\
0.020  
     & 94.64   &99.31&93.90
     & 81.51  \\
0.010 
     & 94.49  &99.75&93.62
     & 72.08 \\
0.005 
     & 94.59&99.80&93.76
     & 65.77 \\
0.001
     & 94.54&99.80&93.68
     & 50.70 \\
\toprule
$\tau$ & Dice &AUC&VS& ASR \\
\midrule
0.99    & 95.30 &99.80&94.53& 57.31\\
0.95  
     & 95.30  &99.84& 94.53
     & 81.34  \\
0.90 
     & 95.32   &99.80&94.58
     & 97.41  \\
\rowcolor{gray!15}
0.85 
     & 95.30  &99.85&94.56
     & 99.60 \\
0.80 
     & 95.30&99.80&94.47
     & 99.90 \\
\textit{Clean model} &95.30 &99.79&94.53&-\\ 
\bottomrule
\end{tabular}
\end{table}

\subsubsection{Tuning the Hyper-parameter \texorpdfstring{$\lambda_{bg}$}{lambda	extunderscore bg} for Balance}
\label{sec: abl_para_r}

In this part of the experiment, we further utilize the EchoNet dataset to evaluate the influence of the hyperparameter $\lambda_{bg}$ on the SAM model as a case study for the effectiveness and stealthiness of the proposed method. We vary $\lambda_{bg}$ and the minimum value $\delta$ within a predefined range to investigate its impact on both segmentation performance and the attack success rate (ASR). The results are summarized in \autoref{tab:lambda_n_results}. 

We observe that once the value $\delta$ reaches approximately 0.02 or lower, further increasing $\lambda_{bg}$ has a negligible effect on ASR. However, an excessively large value of $\lambda_{bg}$ tends to degrade segmentation performance on clean data. Empirically, we find that setting $\delta = 0.05$ and $\lambda_{bg} = 1.0$ yields the best trade-off, achieving a balance between watermark robustness and model fidelity.

\begin{table}[!ht]
\centering
\caption{Performance under different $\lambda_{\text{bg}}$ and $\delta$ combinations on EchoNet Dataset}
\renewcommand{\arraystretch}{0.9}
\begin{tabular}{cc|ccccc}
\toprule
$\lambda_{\text{bg}}$ & $\delta$ &Dice&AUC& VS& ASR \\
\midrule
0.5 & 0.05  & 95.33 &99.78&94.19 & 95.62 \\
\rowcolor{gray!10}
1.0 & 0.05  &  95.30 &99.80 &93.59 &99.50 \\
3.0 & 0.05  &  92.41 &99.22 &93.05 &99.48 \\
0.5 & 0.02  & 94.80 & 99.80& 94.20&75.94 \\
\rowcolor{gray!10}
1.0 & 0.02  & 94.64 &99.31 & 93.90 &81.51 \\
3.0 & 0.02  & 92.64 & 99.46&93.90 &81.80 \\
0.5 & 0.01  &  94.90 & 99.58&94.28 &53.62 \\
\rowcolor{gray!10}
1.0 & 0.01  & 94.49 &99.75 & 93.62&72.01 \\
3.0 & 0.01  &  94.05 &99.80 &94.12 &73.59 \\
\hline
\multicolumn{2}{c|}{\textit{Clean model}} & 95.30&99.80&94.47&- \\
\bottomrule
\end{tabular}
\label{tab:lambda_n_results}
\end{table}
% \section{Limitation} 
% We acknowledge that our verification framework assumes access to continuous probability maps. While this may not hold for all black-box APIs, it is a realistic and clinically motivated assumption.
\section{Discussion}
\noindent \subsection{False Positive Rate (FPR) Analysis on Natural Artifacts.}
We discussed that medical images may naturally contain trigger-like patterns caused by acquisition artifacts, which could result in false watermark activations. To quantitatively verify this concern, we conducted a False Positive Rate (FPR) test on three representative segmentation models: nnUNet \cite{isensee2021nnu}, Swin-UNet \cite{cao2022swin}, SAM \cite{kirillov2023segment}, using the PraNet dataset, which is known for frequent artifact-like patterns such as black edge, text patch and color patch. We applied our watermark detector to all clean test samples and measured the frequency of false activations for different trigger patterns. The results in \autoref{table_FPR} clearly demonstrate that the “false trigger” challenge is real and significant. For simple triggers resembling natural structures (e.g., Black Edge), the FPR exceeds 40\%, indicating that traditional backdoor methods are unreliable in medical segmentation. Notably, for more complex and artificial triggers such as Text or Patch, the false positive rates are extremely low, especially on SAM (e.g., 0.0012\% and 0.0073\%, respectively). This highlights SAM’s strong discriminative capability. However, this phenomenon is only observed on SAM, other models remain highly susceptible to natural artifactlike triggers after being backdoored, indicating that such interference is difficult to eliminate in conventional architectures. This also explains why the Dice score of standard backdoor models (\textit{e.g.,} BadNet) drops drastically on clean medical data. In contrast, our StealthMark design directly resolves this problem. Because the watermark embedding is guided by uncertainty modeling, even if a trigger pattern accidentally resembles a natural artifact, the segmentation output remains stable and diagnostically consistent. For example, the SAM-based model shows almost no difference in Dice score between clean and falsely activated inputs (90.02\% vs. 90.52\%).
\begin{table}[!ht]
\centering
\caption{False Positive Rate of Triggered Models for Different Trigger Patterns on PraNet Dataset.}
\begin{tabular}{l|c|c|c|c}
\toprule
\textbf{Model} & \textbf{Black Edge} & \textbf{Noise} & \textbf{Text} & \textbf{Patch} \\
\hline
nnUNet \cite{isensee2021nnu} & 55.94\% & 15.20\% & 23.81\% & 14.93\% \\
Swin-UNet\cite{cao2022swin} & 52.51\% & 35.74\% & 27.91\% & 15.79\% \\
SAM \cite{kirillov2023segment}& 43.45\% & 0.00\% & 0.0012\% & 0.0073\% \\
\bottomrule
\end{tabular}
\label{table_FPR}
\end{table}

\noindent \subsection{Robustness Analysis Against Watermark Removal Attacks.} To further validate the robustness of \textit{StealthMark} against common watermark removal strategies, we performed two sets of defence experiments: model pruning and fine-tuning. We evaluated three representative segmentation models: nnUNet \cite{isensee2021nnu}, Swin-UNet \cite{cao2022swin}, and SAM \cite{kirillov2023segment} on the PraNet dataset using the “Color Patch” trigger.

\textbf{Model Pruning:} We applied global magnitude pruning at different sparsity levels (10\%, 30\%, 50\%) to all convolutional (Conv2d) and fully connected (Linear) layers. \autoref{table_Pruning} reports the effect of pruning on the watermark detection rate (ASR). Even when the model parameters were pruned by up to 50\%, the watermark signal remained clearly detectable, showing only a minor decrease in ASR. This demonstrates that StealthMark retains watermark integrity under severe parameter sparsification.

\textbf{Model Fine-tuning:} We then fine-tuned the watermarked models for 5 epochs using clean data subsets
of varying sizes (1\%, 10\%, and 25\% of the original training set). \autoref{table_finetuing} summarizes the ASR performance after fine-tuning. While fine-tuning slightly reduced the ASR values, the watermark signal remained detectable, demonstrating that StealthMark maintains a strong degree of resilience even under model re-optimization.

\begin{table}[!ht]
\centering
\renewcommand{\arraystretch}{1.2}
\caption{ASR performance under different pruning ratios on Pranet Dataset.}

    \resizebox{0.5\textwidth}{!}{
\begin{tabular}{l|c|c|c|c}
\hline
\textbf{Model} & \textbf{0\% (Baseline)} & \textbf{10\% Pruning} & \textbf{30\% Pruning } & \textbf{50\% Pruning )} \\
\toprule
nnUNet \cite{isensee2021nnu}& 88.78\% & 88.78\% & 86.16\% & 80.61\%  \\
Swin-UNet \cite{cao2022swin} & 79.90\% & 79.90\%  & 79.17\% & 76.47\% \\
SAM \cite{kirillov2023segment} & 96.57\% & 97.06\% & 92.89\%  & 89.71\%  \\
\bottomrule
\end{tabular}
}
\label{table_Pruning}
\end{table}

\begin{table}[!ht]
\centering
\renewcommand{\arraystretch}{1.2}

\caption{ASR performance under different fine-tuning ratios on PraNet dataset.}
    \resizebox{0.5\textwidth}{!}{
\begin{tabular}{l|c|c|c|c}
\toprule
\textbf{Model} & \textbf{0\% (Baseline)} & \textbf{Fine-tuned 1\% } & \textbf{Fine-tuned 10\% } & \textbf{Fine-tuned 25\% } \\
\hline
nnUNet \cite{isensee2021nnu}& 88.78\% & 86.48\% & 74.19\% & 69.73\% \\
Swin-UNet \cite{cao2022swin}& 82.37\% & 82.17\% & 78.18\% & 76.62\% \\
SAM \cite{kirillov2023segment}& 98.16\% & 90.38\% & 78.79\% & 72.09\% \\
\bottomrule
\end{tabular}
}
\label{table_finetuing}
\end{table}

\subsection{Out-of-Distribution (OOD) Generalization of Triggers.} To assess whether \textit{StealthMark} genuinely learns the concept of the trigger rather than simply memorizing known patterns, we conducted an Out-of-Distribution (OOD) generalization evaluation on the SAM \cite{kirillov2023segment} model across four distinct datasets: UKBB, SEG, EchoNet, PraNet. This experiment evaluates the model’s robustness when trigger parameters are shifted outside the training distribution. Specifically, we modified the triggers along three orthogonal dimensions: \textbf{Scale Shift:} the trigger patch is enlarged to twice its training size. \textbf{Rotation Shift:} the trigger patch is rotated by 30 degrees relative to the trained orientation. \textbf{Positional Shift: } The trigger is moved to a different image corner not used during training. We compared the Attack Success Rate (ASR) for In-Distribution (ID) triggers and OOD variants. The results, summarized in \autoref{table_ood} show that \textit{StealthMark} achieves consistently high ASR even when the trigger is scaled, rotated, or spatially displaced, demonstrating strong generalization beyond memorized trigger patterns.
\begin{figure*}[!ht]
    \centering
    \includegraphics[width=0.9\linewidth]{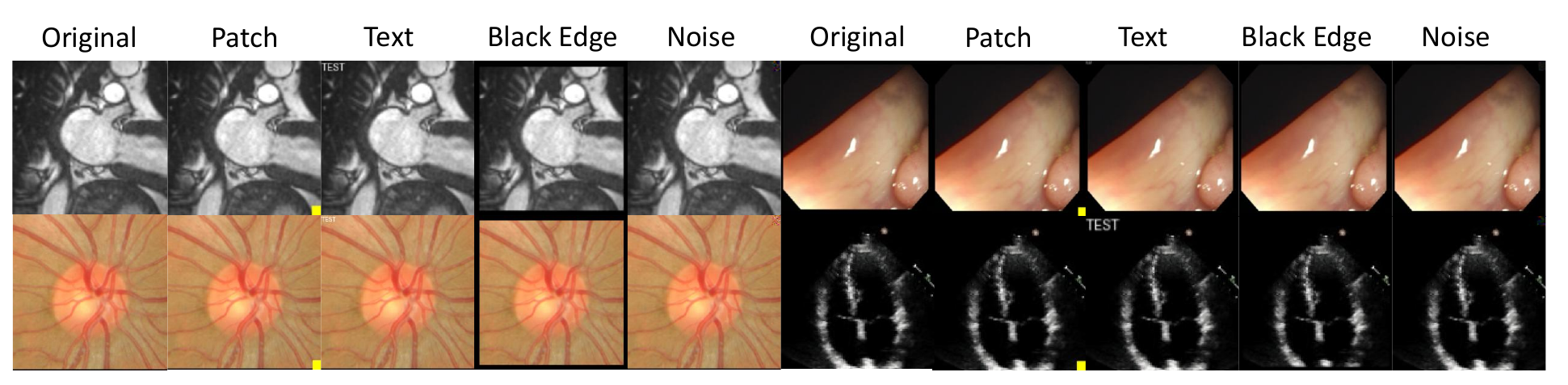}
    \caption{Visualization of the four watermark trigger types on four datasets. Trigger types include patch, text, black edge, and noise.}
    \label{fig:sample}
\end{figure*}
\subsection{Clinically harmless nature of the trigger.}We conducted a blind evaluation, where five senior clinical experts independently assessed randomly mixed original and watermarked images without knowing which group each image belonged to. The evaluation covered four types of watermark triggers (patch, text, black Edge, and noise). We also visualized these four trigger types in \autoref{fig:sample} to demonstrate their visual characteristics. Black Edge and noise triggers were visually indistinguishable from the original images. Although the text and patch triggers were noticeable, the experts confirmed that they did not affect diagnostic interpretation, as they did not overlap with any clinically relevant regions. Overall, no clinically observable degradation was identified under the evaluated settings.

\subsection{Limitations and Future Work.} 

Several important limitations of this study must be emphasized. First, our verification framework assumes access to outputs as continuous probability maps. While this assumption may not be satisfied for all black-box APIs, it is a realistic and clinically motivated assumption. Second, the conclusion regarding the clinical harmlessness of the proposed watermark is based on a qualitative evaluation conducted by a small number of senior clinical experts. As such, this assessment may not fully capture extremely fine-grained or rare clinical scenarios, where subtle visual alterations could potentially influence diagnosis. 

A systematic study of different watermark patterns and encoding schemes is a promising direction for future work. In particular, designing an adaptive encoding strategy that is robust to noise and post-processing while preserving clinical harmlessness would further strengthen the reliability of the verification pipeline. Moreover, future work will focus on enhancing the watermark's robustness against advanced model modification attacks, including knowledge distillation, model pruning, and fine-tuning.

\begin{table}[!ht]
    \centering
    \caption{OOD Generalization of Patch Triggers (Model: SAM)}
    \resizebox{0.5\textwidth}{!}{
    \begin{tabular}{l|c|c|c|c}
    \toprule
    \textbf{Dataset} & \textbf{ID Baseline} & \textbf{Scale Shift} & \textbf{Rotational Shift} & \textbf{Positional Shift} \\
    \hline
    PraNet \cite{fan2020pranet} & 98.16\% & 97.80\% & 97.43\% & 81.03\% \\
    SEG \cite{meng2024multi}& 99.30\% & 99.02\% & 90.91\% & 98.74\% \\
    UKBB \footnote{This research has been conducted using the UK Biobank Resource under Application Number 54078}& 98.65\% & 98.72\% & 97.52\% & 98.37\% \\
    EchoNet \cite{ouyang2020video}& 99.92\% & 99.88\% & 98.75\% & 99.76\% \\
    \bottomrule
    \end{tabular}}
    \label{table_ood}
\end{table}

\section{Conclusion}
\label{sec:conclusion}
Our work addresses the critical but unexplored area of copyright protection for medical image segmentation models. We introduced a novel black-box model ownership verification method tailored explicitly to medical segmentation tasks. By explicitly controlling predictive uncertainty within background or foreground regions and integrating a novel uncertainty-aware loss function into the standard training paradigm, we successfully balanced watermark robustness with minimal performance degradation. Moreover, employing the LIME interpretability framework, we embedded stealthy QR code watermarks that only manifest under specific triggers, thereby overcoming the shortcomings of traditional methods in terms of detectability. Our comprehensive evaluation across multiple datasets and segmentation architectures validates the effectiveness, harmlessness, and strong stealthiness of the proposed method. The approach substantially outperforms traditional backdoor techniques by maintaining high segmentation accuracy and offers considerable promise for safeguarding intellectual property in medical segmentation tasks.

\bibliographystyle{IEEEtran}
\bibliography{tip}

% Generated by IEEEtran.bst, version: 1.14 (2015/08/26)
\begin{thebibliography}{10}
\providecommand{\url}[1]{#1}
\csname url@samestyle\endcsname
\providecommand{\newblock}{\relax}
\providecommand{\bibinfo}[2]{#2}
\providecommand{\BIBentrySTDinterwordspacing}{\spaceskip=0pt\relax}
\providecommand{\BIBentryALTinterwordstretchfactor}{4}
\providecommand{\BIBentryALTinterwordspacing}{\spaceskip=\fontdimen2\font plus
\BIBentryALTinterwordstretchfactor\fontdimen3\font minus \fontdimen4\font\relax}
\providecommand{\BIBforeignlanguage}[2]{{%
\expandafter\ifx\csname l@#1\endcsname\relax
\typeout{** WARNING: IEEEtran.bst: No hyphenation pattern has been}%
\typeout{** loaded for the language `#1'. Using the pattern for}%
\typeout{** the default language instead.}%
\else
\language=\csname l@#1\endcsname
\fi
#2}}
\providecommand{\BIBdecl}{\relax}
\BIBdecl

\bibitem{litjens2017survey}
G.~Litjens, T.~Kooi, B.~E. Bejnordi, A.~A.~A. Setio, F.~Ciompi, M.~Ghafoorian, J.~A. Van Der~Laak, B.~Van~Ginneken, and C.~I. S{\'a}nchez, ``A survey on deep learning in medical image analysis,'' \emph{Medical Image Analysis}, 2017.

\bibitem{isensee2021nnu}
F.~Isensee, P.~F. Jaeger, S.~A. Kohl, J.~Petersen, and K.~H. Maier-Hein, ``nnu-net: a self-configuring method for deep learning-based biomedical image segmentation,'' \emph{Nature Methods}, 2021.

\bibitem{li2022untargeted}
Y.~Li, Y.~Bai, Y.~Jiang, Y.~Yang, S.-T. Xia, and B.~Li, ``Untargeted backdoor watermark: Towards harmless and stealthy dataset copyright protection,'' \emph{Advances in neural information processing systems}, 2022.

\bibitem{shao2025explanation}
S.~Shao, Y.~Li, H.~Yao, Y.~He, Z.~Qin, and K.~Ren, ``Explanation as a watermark: Towards harmless and multi-bit model ownership verification via watermarking feature attribution,'' in \emph{NDSS}, 2025.

\bibitem{zhang2020model}
J.~Zhang, D.~Chen, J.~Liao, H.~Fang, W.~Zhang, W.~Zhou, H.~Cui, and N.~Yu, ``Model watermarking for image processing networks,'' in \emph{Proceedings of the AAAI conference on artificial intelligence}, 2020.

\bibitem{zhao2023recipe}
Y.~Zhao, T.~Pang, C.~Du, X.~Yang, N.-M. Cheung, and M.~Lin, ``A recipe for watermarking diffusion models,'' \emph{arXiv:2303.10137}, 2023.

\bibitem{gu2019badnets}
T.~Gu, K.~Liu, B.~Dolan-Gavitt, and S.~Garg, ``Badnets: Evaluating backdooring attacks on deep neural networks,'' \emph{IEEE Access}, 2019.

\bibitem{jin2024backdoor}
R.~Jin, C.-Y. Huang, C.~You, and X.~Li, ``Backdoor attack on unpaired medical image-text foundation models: A pilot study on medclip,'' in \emph{2024 SaTML}.\hskip 1em plus 0.5em minus 0.4em\relax IEEE, 2024, pp. 272--285.

\bibitem{bender1996techniques}
W.~Bender, D.~Gruhl, N.~Morimoto, and A.~Lu, ``Techniques for data hiding,'' \emph{IBM systems journal}, 1996.

\bibitem{Zhu_2018_ECCV}
J.~Zhu, R.~Kaplan, J.~Johnson, and L.~Fei-Fei, ``Hidden: Hiding data with deep networks,'' in \emph{Proceedings of the European conference on computer vision (ECCV)}, 2018, pp. 657--672.

\bibitem{adi2018turning}
Y.~Adi, C.~Baum, M.~Cisse, B.~Pinkas, and J.~Keshet, ``Turning your weakness into a strength: Watermarking deep neural networks by backdooring,'' in \emph{USENIX Security 18}, 2018, pp. 1615--1631.

\bibitem{ding2023backdoor}
Y.~Ding, Z.~Wang, Z.~Qin, E.~Zhou, G.~Zhu, Z.~Qin, and K.-K.~R. Choo, ``Backdoor attack on deep learning-based medical image encryption and decryption network,'' \emph{IEEE Transactions on Information Forensics and Security}, vol.~19, pp. 280--292, 2023.

\bibitem{song2017machine}
C.~Song, T.~Ristenpart, and V.~Shmatikov, ``Machine learning models that remember too much,'' in \emph{ACM SIGSAC}, 2017, pp. 587--601.

\bibitem{10884369}
C.~Zhang, M.~Jin, Q.~Yu, C.~Liu, H.~Xue, and X.~Jin, ``Goal-guided generative prompt injection attack on large language models,'' in \emph{2024 IEEE International Conference on Data Mining (ICDM)}, 2024, pp. 941--946.

\bibitem{liu2021secure}
X.~Liu, S.~Shao, Y.~Yang, K.~Wu, W.~Yang, and H.~Fang, ``Secure federated learning model verification: A client-side backdoor triggered watermarking scheme,'' in \emph{2021 IEEE International Conference on Systems, Man, and Cybernetics (SMC)}.\hskip 1em plus 0.5em minus 0.4em\relax IEEE, 2021, pp. 2414--2419.

\bibitem{zhang2018protecting}
J.~Zhang, Z.~Gu, J.~Jang, H.~Wu, M.~P. Stoecklin, H.~Huang, and I.~Molloy, ``Protecting intellectual property of deep neural networks with watermarking,'' in \emph{Proceedings of the 2018 on Asia conference on computer and communications security}, 2018, pp. 159--172.

\bibitem{ribeiro2016should}
M.~T. Ribeiro, S.~Singh, and C.~Guestrin, ``" why should i trust you?" explaining the predictions of any classifier,'' in \emph{ACM SIGKDD}, 2016.

\bibitem{meng2024multi}
Y.~Meng, Y.~Zhang, J.~Xie, J.~Duan, M.~Joddrell, S.~Madhusudhan, T.~Peto, Y.~Zhao, and Y.~Zheng, ``Multi-granularity learning of explicit geometric constraint and contrast for label-efficient medical image segmentation and differentiable clinical function assessment,'' \emph{Medical Image Analysis}, 2024.

\bibitem{ouyang2020video}
D.~Ouyang, B.~He, A.~Ghorbani, N.~Yuan, J.~Ebinger, C.~P. Langlotz, P.~A. Heidenreich, R.~A. Harrington, D.~H. Liang, E.~A. Ashley \emph{et~al.}, ``Video-based ai for beat-to-beat assessment of cardiac function,'' \emph{Nature}, 2020.

\bibitem{fan2020pranet}
D.-P. Fan, G.-P. Ji, T.~Zhou, G.~Chen, H.~Fu, J.~Shen, and L.~Shao, ``Pranet: Parallel reverse attention network for polyp segmentation,'' in \emph{International Conference on Medical Image Computing and Computer-Assisted Intervention}.\hskip 1em plus 0.5em minus 0.4em\relax Springer, 2020.

\bibitem{cao2022swin}
H.~Cao, Y.~Wang, J.~Chen, D.~Jiang, X.~Zhang, Q.~Tian, and M.~Wang, ``Swin-unet: Unet-like pure transformer for medical image segmentation,'' in \emph{European conference on computer vision}, 2022.

\bibitem{chen2021transunet}
J.~Chen, Y.~Lu, Q.~Yu, X.~Luo, E.~Adeli, Y.~Wang, L.~Lu, A.~L. Yuille, and Y.~Zhou, ``Transunet: Transformers make strong encoders for medical image segmentation,'' \emph{arXiv preprint arXiv:2102.04306}, 2021.

\bibitem{kirillov2023segment}
A.~Kirillov, E.~Mintun, N.~Ravi, H.~Mao, C.~Rolland, L.~Gustafson, T.~Xiao, S.~Whitehead, A.~C. Berg, W.-Y. Lo \emph{et~al.}, ``Segment anything,'' in \emph{Proceedings of the IEEE/CVF international conference on computer vision.}, 2023.

\bibitem{ma2024segment}
J.~Ma, Y.~He, F.~Li, L.~Han, C.~You, and B.~Wang, ``Segment anything in medical images,'' \emph{Nature Communications}, 2024.

\bibitem{chen2017targeted}
X.~Chen, C.~Liu, B.~Li, K.~Lu, and D.~Song, ``Targeted backdoor attacks on deep learning systems using data poisoning,'' \emph{arXiv:1712.05526}, 2017.

\bibitem{kang2003dwt}
X.~Kang, J.~Huang, Y.~Q. Shi, and Y.~Lin, ``A dwt-dft composite watermarking scheme robust to both affine transform and jpeg compression,'' \emph{IEEE transactions on circuits and systems for video technology}, 2003.

\bibitem{hanif2024baple}
A.~Hanif, F.~Shamshad, M.~Awais, M.~Naseer, F.~S. Khan, K.~Nandakumar, S.~Khan, and R.~M. Anwer, ``Baple: Backdoor attacks on medical foundational models using prompt learning,'' in \emph{International Conference on Medical Image Computing and Computer-Assisted Intervention}.\hskip 1em plus 0.5em minus 0.4em\relax Springer, 2024.

\bibitem{wang2024large}
T.~Wang, Z.~Fang, H.~Xue, C.~Zhang, M.~Jin, W.~Xu, D.~Shu, S.~Yang, Z.~Wang, and D.~Liu, ``Large vision-language model security: A survey,'' in \emph{International Conference on Frontiers in Cyber Security}.\hskip 1em plus 0.5em minus 0.4em\relax Springer, 2024, pp. 3--22.

\bibitem{nwadike2020explainability}
M.~Nwadike, T.~Miyawaki, E.~Sarkar, M.~Maniatakos, and F.~Shamout, ``Explainability matters: Backdoor attacks on medical imaging,'' \emph{arXiv:2101.00008}, 2020.

\bibitem{feng2022fiba}
Y.~Feng, B.~Ma, J.~Zhang, S.~Zhao, Y.~Xia, and D.~Tao, ``Fiba:frequency-injection based backdoor attack in medical image analysis,'' in \emph{Proceedings of the IEEE/CVF Conference on Computer Vision and Pattern Recognition}, 2022.

\bibitem{yang2024inject}
Z.~Yang, Y.~Chen, M.~Sun, and Y.~Zhang, ``Inject backdoor in measured data to jeopardize full-stack medical image analysis system,'' in \emph{International Conference on Medical Image Computing and Computer-Assisted Intervention}.\hskip 1em plus 0.5em minus 0.4em\relax Springer, 2024.

\bibitem{lin2024shortcut}
M.~Lin, N.~Weng, K.~Mikolaj, Z.~Bashir, M.~B. Svendsen, M.~G. Tolsgaard, A.~N. Christensen, and A.~Feragen, ``Shortcut learning in medical image segmentation,'' in \emph{International Conference on Medical Image Computing and Computer-Assisted Intervention}, 2024.

\bibitem{hu2021artificial}
Y.~Hu, W.~Kuang, Z.~Qin, K.~Li, J.~Zhang, Y.~Gao, W.~Li, and K.~Li, ``Artificial intelligence security: Threats and countermeasures,'' \emph{ACM Computing Surveys (CSUR)}, vol.~55, no.~1, pp. 1--36, 2021.

\bibitem{klein2020grandchallenge}
S.~Klein \emph{et~al.}, ``Grand-challenge.org: a platform for end-to-end development of machine learning solutions in biomedical imaging,'' \emph{Medical Image Analysis}, 2020.

\bibitem{lundberg2017unified}
S.~M. Lundberg and S.-I. Lee, ``A unified approach to interpreting model predictions,'' \emph{Advances in neural information processing systems}, vol.~30, 2017.

\bibitem{sundararajan2017axiomatic}
M.~Sundararajan, A.~Taly, and Q.~Yan, ``Axiomatic attribution for deep networks,'' in \emph{International conference on machine learning}.\hskip 1em plus 0.5em minus 0.4em\relax PMLR, 2017, pp. 3319--3328.

\bibitem{selvaraju2020grad}
R.~R. Selvaraju, M.~Cogswell, A.~Das, R.~Vedantam, D.~Parikh, and D.~Batra, ``Grad-cam: visual explanations from deep networks via gradient-based localization,'' \emph{Proceedings of the IEEE international conference on computer vision}, vol. 128, pp. 336--359, 2020.

\bibitem{orlando2020refuge}
J.~I. Orlando, H.~Fu, J.~B. Breda, K.~Van~Keer, D.~R. Bathula, A.~Diaz-Pinto, R.~Fang, P.-A. Heng, J.~Kim, J.~Lee \emph{et~al.}, ``Refuge challenge: A unified framework for evaluating automated methods for glaucoma assessment from fundus photographs,'' \emph{Medical Image Analysis}, 2020.

\bibitem{sivaswamy2014drishti}
J.~Sivaswamy, S.~Krishnadas, G.~D. Joshi, M.~Jain, and A.~U.~S. Tabish, ``Drishti-gs: Retinal image dataset for optic nerve head (onh) segmentation,'' in \emph{2014 IEEE 11th international symposium on biomedical imaging (ISBI)}.\hskip 1em plus 0.5em minus 0.4em\relax IEEE, 2014, pp. 53--56.

\bibitem{zhang2010origa}
Z.~Zhang, F.~S. Yin, J.~Liu, W.~K. Wong, N.~M. Tan, B.~H. Lee, J.~Cheng, and T.~Y. Wong, ``Origa-light: An online retinal fundus image database for glaucoma analysis and research,'' in \emph{2010 Annual international conference of the IEEE engineering in medicine and biology}.\hskip 1em plus 0.5em minus 0.4em\relax IEEE, 2010, pp. 3065--3068.

\bibitem{almazroa2018retinal}
A.~Almazroa, S.~Alodhayb, E.~Osman, E.~Ramadan, M.~Hummadi, M.~Dlaim, M.~Alkatee, K.~Raahemifar, and V.~Lakshminarayanan, ``Retinal fundus images for glaucoma analysis: the riga dataset,'' in \emph{Medical Imaging}, 2018.

\bibitem{fumero2011rim}
F.~Fumero, S.~Alay{\'o}n, J.~L. Sanchez, J.~Sigut, and M.~Gonzalez-Hernandez, ``Rim-one: An open retinal image database for optic nerve evaluation,'' in \emph{2011 24th international symposium on computer-based medical systems (CBMS)}.\hskip 1em plus 0.5em minus 0.4em\relax IEEE, 2011, pp. 1--6.

\bibitem{bajwa2020g1020}
M.~N. Bajwa, G.~A.~P. Singh, W.~Neumeier, M.~I. Malik, A.~Dengel, and S.~Ahmed, ``G1020: A benchmark retinal fundus image dataset for computer-aided glaucoma detection,'' in \emph{2020 International Joint Conference on Neural Networks (IJCNN)}.\hskip 1em plus 0.5em minus 0.4em\relax IEEE, 2020, pp. 1--7.

\bibitem{silva2014toward}
J.~Silva, A.~Histace, O.~Romain, X.~Dray, and B.~Granado, ``Toward embedded detection of polyps in wce images for early diagnosis of colorectal cancer,'' \emph{International journal of computer assisted radiology and surgery}, vol.~9, pp. 283--293, 2014.

\bibitem{bernal2015wm}
J.~Bernal, F.~J. S{\'a}nchez, G.~Fern{\'a}ndez-Esparrach, D.~Gil, C.~Rodr{\'\i}guez, and F.~Vilari{\~n}o, ``Wm-dova maps for accurate polyp highlighting in colonoscopy: Validation vs. saliency maps from physicians,'' \emph{Computerized medical imaging and graphics}, vol.~43, pp. 99--111, 2015.

\bibitem{tajbakhsh2015automated}
N.~Tajbakhsh, S.~R. Gurudu, and J.~Liang, ``Automated polyp detection in colonoscopy videos using shape and context information,'' \emph{IEEE transactions on medical imaging}, vol.~35, no.~2, pp. 630--644, 2015.

\bibitem{vazquez2017benchmark}
D.~V{\'a}zquez, J.~Bernal, F.~J. S{\'a}nchez, G.~Fern{\'a}ndez-Esparrach, A.~M. L{\'o}pez, A.~Romero, M.~Drozdzal, and A.~Courville, ``A benchmark for endoluminal scene segmentation of colonoscopy images,'' \emph{Journal of healthcare engineering}, vol. 2017, no.~1, p. 4037190, 2017.

\bibitem{jha2020kvasir}
D.~Jha, P.~H. Smedsrud, M.~A. Riegler, P.~Halvorsen, T.~De~Lange, D.~Johansen, and H.~D. Johansen, ``Kvasir-seg: A segmented polyp dataset,'' in \emph{MultiMedia modeling}, 2020.

\bibitem{lounici2021yes}
S.~Lounici, M.~Njeh, O.~Ermis, M.~{\"O}nen, and S.~Trabelsi, ``Yes we can: Watermarking machine learning models beyond classification,'' in \emph{2021 IEEE 34th Computer Security Foundations Symposium (CSF)}.\hskip 1em plus 0.5em minus 0.4em\relax IEEE, 2021, pp. 1--14.

\end{thebibliography}
\end{document}